%% file: main.tex
\documentclass[sigconf,authorversion,screen]{acmart}
\AtBeginDocument{%
  }

\copyrightyear{2026}
\acmYear{2026}
\setcopyright{cc}
\setcctype{by}
\acmConference[MM '26]{Proceedings of the 34th ACM International Conference on Multimedia}{November 10--14, 2026}{Rio de Janeiro, Brazil}
\acmBooktitle{Proceedings of the 34th ACM International Conference on Multimedia (MM '26), November 10--14, 2026, Rio de Janeiro, Brazil}
\acmDOI{10.1145/3767308.3835237}
\acmISBN{979-8-4007-2213-4/2026/11}

\usepackage{colortbl}

\usepackage{multirow} 

\newcommand{\eg}{\emph{e.g.}}
\newcommand{\ie}{\emph{i.e.}}

\definecolor{Dred}{RGB}{220,20,1}
\definecolor{Dgreen}{RGB}{26,153,25}
\definecolor{gray}{HTML}{F2F2F2}
\definecolor{Lblue}{HTML}{e3f1ff} 
\definecolor{Mblue}{HTML}{f7faff} 
\definecolor{Dblue}{HTML}{005ed3} 
\definecolor{GTblue}{HTML}{005ed3}
\definecolor{LTgray}{RGB}{140,137,154}

\usepackage{pifont}     

\newcommand{\cmark}{\ding{51}}

\newcommand{\up}{$\uparrow$}
\newcommand{\down}{$\downarrow$}

\newcommand{\circnum}[1]{\ding{\numexpr171+#1\relax}}
\newcommand{\blackcircnum}[1]{\ding{\numexpr181+#1\relax}}

\newcommand{\para}[1]{\textbf{#1}.}

\newcommand{\first}[1]{\textbf{#1}}
\newcommand{\second}[1]{\underline{#1}}

\newcommand{\gt}[1]{\textcolor{GTblue}{\fontsize{7}{8}\selectfont (+{#1})}}
\newcommand{\gtsign}[1]{\textcolor{GTblue}{\fontsize{7}{8}\selectfont{#1}}}

\newcommand{\ls}[1]{\textcolor{LTgray}{\fontsize{7}{8}\selectfont (-{#1})}}
\newcommand{\lssign}[1]{\textcolor{LTgray}{\fontsize{7}{8}\selectfont{#1}}}

\newcommand{\name}[1]{\textcolor{black}{#1}}

\newcommand{\LAB}{AdaLAB}
\newcommand{\FRAMEWORK}{CAGE}
\newcommand{\TRANSFORM}{AdaCCT}

\newcommand{\ours}{\rowcolor{Lblue}}
\newcommand{\oursDelta}{\rowcolor{Mblue}}

\newcommand{\oursrow}{\ours \quad +Ours}
\newcommand{\deltarow}{\oursDelta \quad $\Delta$}

\begin{document}

\title{Towards Color-Faithful Low-Light Image Enhancement via Adaptive Color Debiasing and Saturation Rectification}

\author{\name{Zhichen Yang}}
\orcid{0009-0000-6204-7718}
\affiliation{%
  \institution{Fuzhou University}
  \city{Fuzhou}
  \state{Fujian}
  \country{China}
}
\email{zhichenyang47@gmail.com}

\author{\name{Rui Xu}}
\orcid{0000-0003-3767-6816}
\affiliation{%
  \institution{Fuzhou University}
  \city{Fuzhou}
  \state{Fujian}
  \country{China}
}
\email{xurui.ryan.chn@gmail.com}

\author{\name{Yuzhen Niu}}
\orcid{0000-0002-9874-9719}
\authornote{Corresponding Author: yuzhenniu@gmail.com}
\affiliation{%
  \institution{Fuzhou University}
  \city{Fuzhou}
  \state{Fujian}
  \country{China}
}
\email{yuzhenniu@gmail.com}

\author{\name{Fusheng Li}}
\orcid{0009-0002-3851-9118}
\affiliation{%
  \institution{Fuzhou University}
  \city{Fuzhou}
  \state{Fujian}
  \country{China}
}
\email{lifusheng.chn@gmail.com}

\author{\name{Hui Da}}
\orcid{0009-0006-9503-7835}
\affiliation{%
  \institution{Fuzhou University}
  \city{Fuzhou}
  \state{Fujian}
  \country{China}
}
\email{dahuifzu@gmail.com}

\author{\name{Ri Cheng}}
\orcid{0000-0002-5866-6847}
\affiliation{%
  \institution{Fuzhou University}
  \city{Fuzhou}
  \state{Fujian}
  \country{China}
}
\email{rcheng@fzu.edu.cn}

\renewcommand{\shortauthors}{Yang et al.}


\begin{abstract}
Low-light imaging often introduces color bias caused by the low signal-to-noise ratio and the image formation process. Although recent low-light image enhancement methods have achieved strong brightness recovery, faithful color restoration remains challenging, manifesting as overall color bias together with local under- and over-saturation.
To address this issue, we propose CAGE, a cylindrical color correction framework with adaptive color debiasing and gamut-harmonized saturation rectification for color-faithful low-light image enhancement. We first introduce AdaLAB, a cylindrical adaptive LAB color space that provides a decoupled and image-specific basis for uniform color correction. Building on this color space, we further develop AdaCCT, an adaptive cylindrical color transform with forward and inverse transforms for the conversion between RGB and AdaLAB color space, as well as necessary color debiasing and saturation rectification. The forward transform suppresses embedded color bias before backbone enhancement by reorganizing the chromatic distribution through chromatic-plane shifting and scaling, while the inverse transform achieves faithful saturation rectification through out-of-gamut lightness compensation. Extensive experiments on multiple benchmarks show that CAGE achieves more faithful color restoration, specifically reduces color bias and saturation abnormality, and delivers better overall visual quality across different low-light enhancement backbones. The code is available at \url{https://yangzhichen763.github.io/CAGE/}.
\end{abstract}


\begin{teaserfigure}
  \centering
  \vspace{-2mm}
  \includegraphics[width=\textwidth]{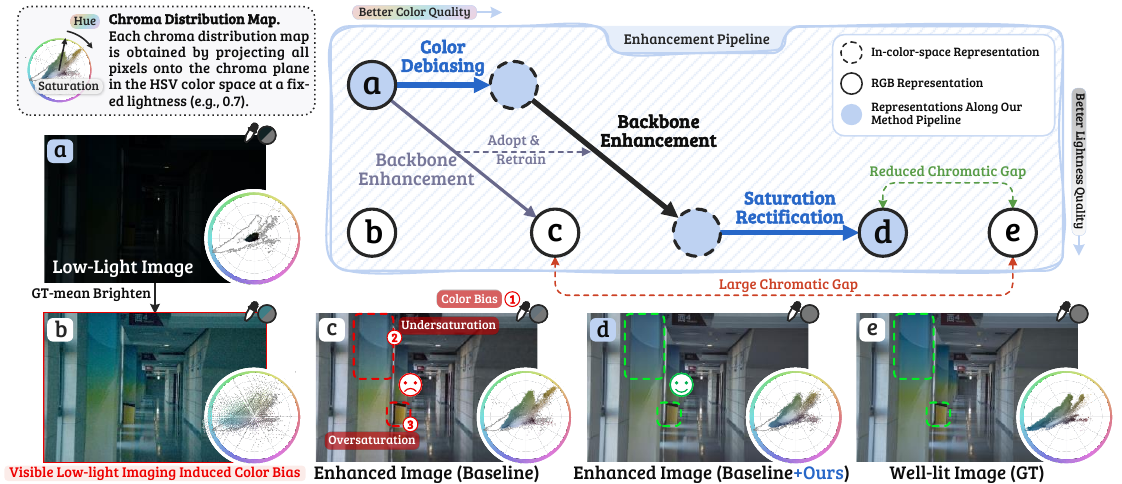}
  \vspace{-4mm}
  \caption{Motivation of our method.
  \emph{(a$\rightarrow$b)} Direct brightening makes the color shift pattern inside low-light images more visible.
  \emph{(a$\rightarrow$c)} Baseline enhancement improves brightness but still leaves a large chromatic gap from the well-lit image, with noticeable color bias, local under- and over-saturation.
  \emph{(a$\rightarrow$d)} Our method improves color faithfulness by adaptive color debiasing before backbone enhancement and gamut-harmonized saturation rectification after enhancement.}
  \Description{
  The figure compares a low-light input, a directly brightened image, a baseline-enhanced image, the CAGE-enhanced image, and the well-lit target. The baseline result contains global color bias together with local under-saturation and over-saturation, whereas CAGE reduces the chromatic gap from the well-lit target.
  }
  \label{fig:teaser}
\end{teaserfigure}


\maketitle



\section{Introduction}

Images captured under low-light conditions often suffer from degraded visibility, insufficient brightness, and distorted colors~\cite{survey-tpami2022}. Low-light image enhancement (LLIE) aims to restore visually plausible brightness and faithful colors from these degraded observations. Although recent methods~\cite{LL.GLARE, LL.BiFormer, LL.UHDFour} have made steady progress in brightness enhancement, faithful color restoration remains difficult. In low-light imaging, the low signal-to-noise ratio, camera hardware limitations, and automatic in-camera processing often introduce noticeable color shift during image formation~\cite{Low-light-Imaging}. 
Once the image is brightened, as shown in Fig.~\ref{fig:teaser}\emph{(b)}, this embedded color shift becomes more visible, making it hard to restore brightness and faithful color simultaneously.

Most existing LLIE methods~\cite{LL.LLFormer, IR.URWKV, IR.Restormer} restore images directly in the RGB space. In this color space, brightness enhancement and color restoration are tightly coupled, making stronger enhancement lead to more severe color distortion~\cite{LL.HVI-CIDNet}, \eg, color shift and saturation abnormality. To ease this coupling, recent methods decouple brightening and color restoration through Retinex-based decomposition~\cite{LL.RetinexNet.Dataset.LOLv1, LL.URetinexNet, LL.Retinexformer, LL.IRetinex} or HSV-inspired color spaces (\eg, HSV~\cite{LL.HSV} and HVI~\cite{LL.HVI-CIDNet}). Although these designs separate lightness and chrominance (\ie, color components independent of lightness), they do not remove the chromatic disturbance already embedded in the low-light input. Instead, this disturbance keeps propagating through the enhancement pipeline and gradually shifts the recovered color distribution away from that of a well-lit image. As illustrated in Fig.~\ref{fig:teaser}\emph{(c)}, the enhanced image therefore exhibits an overall color bias, accompanied by local under- and over-saturation.

To further decouple brightness enhancement from color restoration,
another line of work revisits LLIE in color spaces such as YUV~\cite{LL.LYT-Net, LL.BreaD} and Lab-like spaces~\cite{LL.CWAN}. These spaces not only decouple lightness and chrominance, more importantly, they also provide a more coherent basis for organizing color shift patterns. In particular, the mapping from RGB to Lab-like spaces preserves the consistency in chromatic direction and continuity of chromatic variation, allowing colors to be reorganized in a more structured form. This advantage is especially beneficial for low-light image color restoration, where the key difficulty lies not only in decoupling lightness and chrominance, but also in reorganizing disturbed colors toward well-lit distribution.
Motivated by this, we revisit color-faithful LLIE in CIELab~\cite{CIELab} (denoted by LAB hereafter), whose coherent lightness-chrominance structure offers a more suitable foundation for modeling and correcting low-light-induced color bias.

However, any color space alone is not enough to remove the chromatic disturbance induced by low-light imaging. As an appropriate color space can reorganize distorted colors, but cannot by itself \emph{\blackcircnum{1} correct the embedded color bias} and \emph{\blackcircnum{2} rectify the abnormal saturation caused by the enhancement process}. For the former, the color shift embedded in the low-light image continues to interfere with subsequent restoration and persists even after a color space transformation, leading to biased color restoration (\emph{cf.} Fig.~\ref{fig:teaser}\emph{(c)}, \circnum{1}). For the latter, the enhancement model tends to amplify chromatic responses in a content-dependent manner, which typically results in the coexistence of under-saturation in some regions and over-saturation in others (\emph{cf.} Fig.~\ref{fig:teaser}\emph{(c)}, \circnum{2} and \circnum{3}). Without an explicit mechanism to regulate these behaviors, the chromatic distribution remains inconsistent, and the gap between the enhanced and well-lit images cannot be effectively reduced.

To address these issues, we propose CAGE, a cylindrical \textbf{C}olor correction framework with \textbf{A}daptive color debiasing and \textbf{G}amut-harmonized saturation rectification for low-light image \textbf{E}nhancement.
Specifically, we first construct a cylindrical adaptive LAB \emph{color space}, named AdaLAB, where L and AB inherit the lightness and the two opponent chromatic axes from CIELab respectively, serving as a decoupled and image-specific basis for accommodating uniform color shifts and scaling in the chromatic plane.
Built on this color space, we further develop an adaptive cylindrical \emph{color transform}~(AdaCCT), which learns hue-shift magnitude and chroma-scaling intensity, and adaptively predicts hue-shift direction and lightness sensitivity to reorganize the color distribution before and after backbone enhancement. In the forward transform, AdaCCT suppresses the color bias embedded in the low-light input before backbone enhancement by reorganizing the chromatic distribution. In the inverse transform, AdaCCT rectifies saturation abnormality after backbone enhancement by converting out-of-gamut chroma surplus into lightness compensation rather than direct clipping. With this design, CAGE is compatible with different LLIE backbones and performs more faithful color restoration.

Our contributions are summarized as follows:
\begin{itemize}
    \item We revisit low-light image enhancement (LLIE) from the perspective of embedded chromatic disturbance and reveal that decoupling lightness and chrominance alone cannot eliminate this disturbance, which persists through enhancement and appears as global color bias and saturation abnormality.

    \item We introduce AdaLAB and AdaCCT as a unified color correction design for LLIE. AdaLAB provides a coherent basis for organizing low-light chromatic variation. Built on AdaLAB, AdaCCT suppresses low-light color bias through adaptive hue shifting and chroma scaling before backbone enhancement, and rectifies post-enhancement saturation abnormality through out-of-gamut lightness compensation.

    \item Extensive experiments on both standard and more challenging benchmarks show that CAGE consistently improves color fidelity, saturation naturalness, and overall visual quality across different low-light image enhancement backbones.
\end{itemize}

\section{Related Work}

\begin{figure*}[h]
  \centering
  \includegraphics[width=\linewidth]{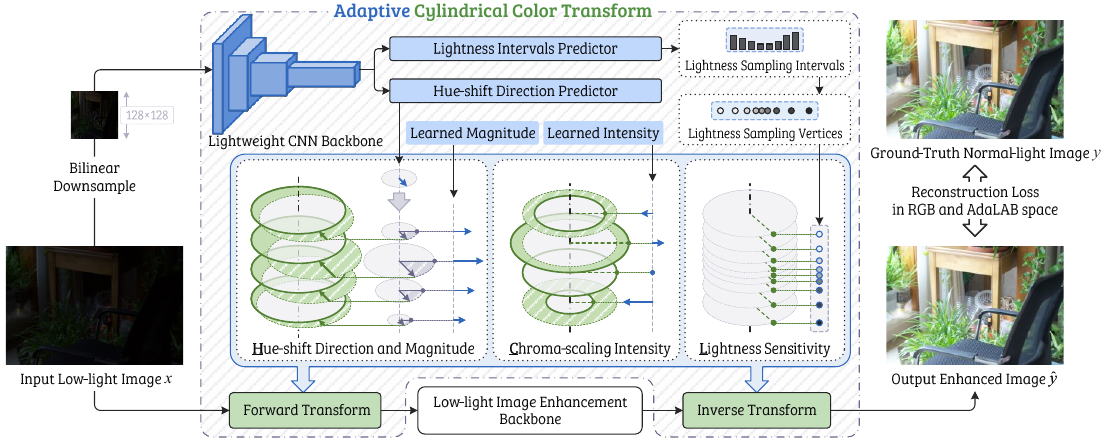}
  \caption{Overview of our proposed framework with a plug-and-play adaptive cylindrical color transform built on the AdaLAB space. AdaCCT predicts image-adaptive cylindrical parameters for the forward and inverse transforms, reducing color bias before backbone enhancement and rectifying saturation abnormality after backbone enhancement.} 
  \Description{
  The framework predicts lightness sensitivity vertices, chroma-scaling intensities, and hue-shift parameters from the downsampled low-light image. The forward transform maps the input into AdaLAB before backbone enhancement, and the inverse transform reconstructs the RGB output with out-of-gamut lightness compensation.
  }
  \vspace{2mm}
  \label{fig:framework}
\end{figure*}

\subsection{Low-light Image Enhancement}

Existing low-light image enhancement methods mainly improve images either in a unified representation~\cite{LL.LLFormer, IR.Restormer, LL.DarkIR, IR.URWKV} or in a decomposed manner, especially in Retinex-based frameworks~\cite{LL.RetinexNet.Dataset.LOLv1, LL.Retinexformer, LL.Diff-Retinex++, LL.LightenDiffusion}. It benefits brightness recovery and structure preservation by separating illumination estimation and color correction from scene reconstruction. However, although decomposition weakens the coupling between illumination and reflectance, it does not explicitly correct the color bias embedded in low-light inputs. The embedded color bias is therefore propagated through enhancement, where amplified chromatic responses can disturb the balance between brightness and color and cause local under- or over-saturation.

To further separate brightness from chromatic content, recent methods revisit low-light enhancement in transformed spaces, but these spaces present different limitations. YUV~\cite{LL.LYT-Net, LL.BreaD, LL.LCNet} offers a linear separation between lightness and chrominance, yielding a perceptually meaningful luminance component but insufficiently separating the two components. HSV~\cite{LL.HSV, LL.BCNet} provides stronger separation, but it suffers from structural artifacts, such as hue discontinuity and black-plane noise~\cite{LL.HVI-CIDNet+}, which destabilize the enhancement results. HVI~\cite{LL.HVI-CIDNet, LL.HVI-CIDNet+} is built on HSV, whose non-differentiable transformation makes color trajectories less smooth and less consistent under the same color bias. CIELab~\cite{LL.CWAN} instead offers smoother chromatic trajectories and a more coherent organization of low-light color shifts. However, a symmetric reversible transform only maps the biased chromatic distribution into a different coordinate system. It neither determines the correction direction nor adjusts the distortion magnitude, so the embedded color bias remains unresolved after the transformation.

\subsection{Color Transform}

Color transforms are widely used in image enhancement for their compact and controllable color mapping. Representative forms include tone curves~\cite{LL.Zero-DCE, LL.ZeroDCE++, CT.NamedCurves}, white balance~\cite{CT.TimeAWB}, and lookup-table-based transforms~\cite{CT.3DLUT, CT.AdaInt, CT.SepLUT, CT.BPAM, CT.SVDLUT}. Among them, image-adaptive 3D look-up tables (3D LUTs)~\cite{CT.3DLUT} enable content-aware color mapping, and later works further extend this formulation with adaptive sampling~\cite{CT.AdaInt}, separable decomposition~\cite{CT.SepLUT}, pixel-wise local adjustment~\cite{CT.BPAM}, and spatial-aware parameterization~\cite{CT.SVDLUT}. However, these methods are mainly designed for general image adjustment rather than low-light-specific color restoration.

Recently, color transforms have also been introduced into low-light enhancement pipelines~\cite{LL.BreaD, LL.HVI-CIDNet, LL.HVI-CIDNet+}. In this setting, they are integrated into decoupled representations to facilitate brightness adjustment and chromatic correction. For example, BreaD~\cite{LL.BreaD} adopts YCbCr decoupling for illumination-dominated color correction, while HVI-CIDNet~\cite{LL.HVI-CIDNet} builds a learnable transform in the HVI space. Even so, existing methods still treat color transform mainly as a supporting component, rather than a dedicated mechanism for suppressing embedded color distortion during reconstruction.

\section{Methodology}
In this section, we first give the overview of CAGE (in Sec.~\ref{sec:overview}). We then describe the image-adaptive cylindrical parameters predicted for the AdaCCT (in Sec.~\ref{sec:parameters}). Finally, we present the forward and inverse transforms of AdaCCT and AdaLAB color space (in Sec.~\ref{sec:adacct}).

\subsection{Overview}
\label{sec:overview}
Following previous work~\cite{LL.HVI-CIDNet}, CAGE integrates a low-light image enhancement backbone with a forward color transform and an inverse color transform of AdaCCT, which can be formulated as:
\begin{equation}
\hat{y} = G^{-1}\!\left(F_\theta\!\left(G(x)\right)\right),
\end{equation}
where $x\in\mathbb{R}^{3\times H\times W}$ and $\hat{y}\in\mathbb{R}^{3\times H\times W}$ denote the input low-light image and the enhanced output image, $F_\theta$ denotes the enhancement backbone, and $G(\cdot)$ and $G^{-1}(\cdot)$ denote the forward and inverse transforms, respectively, which map the image between color spaces. Beyond simple color-space mapping, CAGE further introduces image-adaptive cylindrical parameters in the transformed color space and incorporates color correction into both the forward and inverse transforms to explicitly regulate the chromatic distribution throughout the enhancement pipeline.

Specifically, as illustrated in Fig.~\ref{fig:framework}, given an input image $x$, CAGE first predicts image-adaptive cylindrical parameters from the downsampled input image, including lightness sensitivity vertices, chroma-scaling intensity values, and hue-shift direction vector and magnitude values. Based on these parameters, the forward transform maps the input image from RGB to the proposed AdaLAB color space and suppresses the embedded color bias before backbone enhancement. The enhancement backbone then restores visibility and structures in the color space transformed domain. Finally, the inverse transform maps the enhanced result back to RGB while rectifying out-of-gamut saturation.

\subsection{Image-Adaptive Cylindrical Parameters}
\label{sec:parameters}

Low-light images are formed under varying illumination conditions, camera pipelines, and device responses, where the embedded chromatic disturbance typically does not follow a fixed pattern across images. To make the transform responsive to each input image and its lightness distribution, we predict all image-adaptive parameters from a downsampled input image $x_{\downarrow128}\in\mathbb{R}^{3\times128\times128}$, which reduces the computational overhead. Specifically, a lightweight backbone extracts a compact image vector representation $z\in\mathbb{R}^{D}$:
\begin{equation}
z = H(x_{\downarrow128}),
\end{equation}
where $H(\cdot)$ denotes the lightweight CNN model. Since LAB separates lightness from chromatic plane, and color correction in the chromatic plane can be expressed by directional shift and radial scaling, we then predict three groups of image-adaptive parameters based on $z$ to parameterize a LAB-suited cylindrical color correction, including lightness sensitivity vertices, chroma-scaling intensity values, and hue-shift direction vector and magnitude values.

\para{Lightness Sensitivity Vertices}
Human perception is more sensitive to chromatic variation around moderate lightness levels~\cite{Color}. Natural images generally exhibit a non-uniform distribution along the lightness axis, with pixel values concentrated more heavily in some lightness ranges than others. In low-light enhancement, both embedded color bias and saturation abnormality after enhancement vary with lightness. Therefore, a uniform parameterization on the lightness axis is suboptimal and lacks sufficient flexibility to capture this non-uniformity. To adapt the transform to this non-uniformity, inspired by adaptive interval learning~\cite{CT.AdaInt}, we predict image-adaptive lightness intervals from the input image and convert them into monotonically increasing sampling vertices.

Specifically, given the compact representation $z$, we use a single linear layer as a lightness intervals predictor to predict $p$ interval logits $\{\hat{v}_k\}_{k=1}^{p}$, where $\hat{v}_k\in\mathbb{R}$. These logits are normalized into non-negative interval weights summing to $1$, and accumulated to obtain the lightness sampling vertices, which can be formulated as:
\begin{subequations}
\label{eq:lightness_sampling_vertices}
\begin{align}
\tilde{v}_k &= \frac{\exp(\hat{v}_k)}{\sum_{j=1}^{p}\exp(\hat{v}_j)},
\label{eq:lightness_sampling_vertices-1} \\
v_0 &= 0,\quad
v_k = v_{k-1} + \tilde{v}_k,\quad k=1,\dots,p .
\label{eq:lightness_sampling_vertices-2}
\end{align}
\end{subequations}
where $p$ denotes the number of lightness intervals, and $p+1$ denotes the number of lightness sensitivity vertices.

\para{Chroma-Scaling Intensity}
Low-light imaging often introduces color distortion that varies with lightness~\cite{Dataset.LL.SID}, causing pixels at different lightness levels to drift toward different chromatic ranges (\ie, misalignment of chromatic distributions). A single monotonic scaling curve~\cite{LL.HVI-CIDNet} can reduce this mismatch, but it is not suitable for LAB, where the feasible chroma range is non-monotonic along the lightness axis. A 1D LUT~\cite{CT.1DLUT} along the lightness axis provides greater flexibility, but is less well suited to low-light image enhancement, since each lightness value depends on only two adjacent samples, leaving sparsely populated high-lightness regions weakly supervised and causing scaling inconsistency across intervals.

To better align chromatic distributions across different lightness levels to obtain structured representations in the LAB space, we introduce a group of learned chroma-scaling intensity values with global interaction across all vertices on the lightness axis:
\begin{subequations}
\label{eq:global_interaction}
\begin{align}
\tilde{c}_k &= \sum_{j=0}^{p} \hat{c}_j e^{-\tau |j-k|},
\label{eq:global_interaction-1} \\
c_k &= \operatorname{softplus}\!\left(\tilde{c}_k + 1\right),
\label{eq:global_interaction-2}
\end{align}
\end{subequations}
where $\{\hat{c}_k\}_{k=0}^{p}$ denote the learnable scalar parameters with $\hat{c}_k \in \mathbb{R}$, $e^{-\tau |j-k|}$ is an exponentially decaying weight over the index distance $|j-k|$, and $\tau$ is a learnable temperature controlling the decay rate. $\operatorname{softplus}(\cdot)$ is a smooth activation function that keeps $c_k$ positive, and the additional constant $1$  moves $c_k$ toward unit scaling.

\begin{figure}[t]
  \centering
  \includegraphics[width=\linewidth]{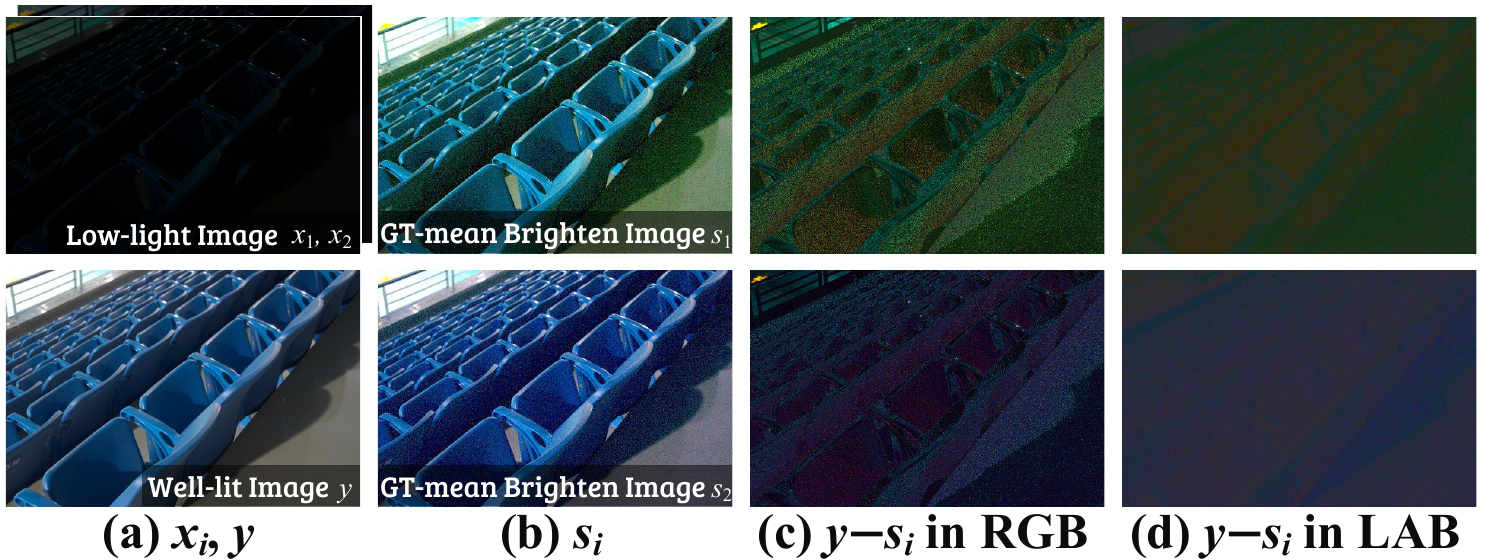}
  \caption{Visualization of low-light color shift patterns. The residuals become more coherent in the proposed AdaLAB.}
  \Description{
  Two low-light examples and their well-lit targets are used to visualize color residuals. The residual maps are scattered and spatially inconsistent in the RGB space, while the corresponding residuals form more compact and directionally consistent patterns in the LAB space.
  }
  \label{fig:color-shift}
\end{figure}

\begin{figure*}[h]
  \centering
  \includegraphics[width=\linewidth]{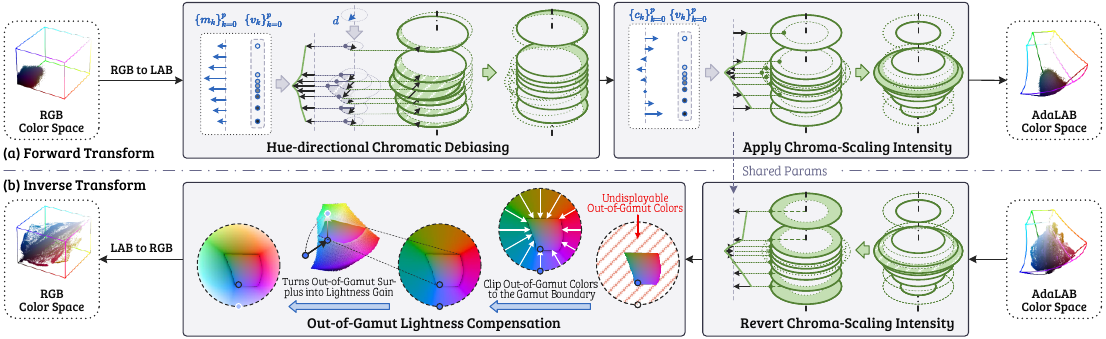}
  \caption{Pipeline of the forward and inverse transforms of the proposed AdaCCT. (a) The forward transform maps the input RGB image to LAB and constructs the AdaLAB representation through hue-directional chromatic debiasing and lightness-aware chroma scaling. (b) The inverse transform reconstructs the enhanced RGB image by reverting the chroma-scaling intensity and applying out-of-gamut lightness compensation before LAB-to-RGB conversion.}
  \Description{
  The forward transform converts an RGB image to LAB, applies hue-directional chromatic debiasing and lightness-aware chroma scaling, and produces an AdaLAB representation. The inverse transform reverts the chroma scaling, converts out-of-gamut chroma surplus into lightness compensation, and maps the corrected LAB representation back to RGB.
  }
  \label{fig:transform}
\end{figure*}

\para{Hue-shift Direction and Magnitude}
Within any single low-light image, color bias typically appears as a globally coherent shift pattern rather than completely irregular pixel-wise variation~\cite{CT.TimeAWB}, which is induced by the low-light imaging process~\cite{Low-light-Imaging}. As shown in Fig.~\ref{fig:color-shift}, the color difference exhibits a more compact pattern in LAB. Since the LAB color space separates lightness from the chromatic plane, this bias pattern can be described more directly by a shift direction together with a lightness-aware shift magnitude.

Specifically, given the compact representation $z$, we use a linear layer as a hue-shift direction predictor $\phi_{\mathrm{d}}$ to predict a shared two-dimensional hue-shift direction vector $d = \phi_{\mathrm{d}}(z)\in\mathbb{R}^{2}$ to model the global shift pattern. Meanwhile, we obtain hue-shift magnitudes $\{m_k\}_{k=0}^{p}$ through the same global interaction as in Eq.~\eqref{eq:global_interaction}, and combine them with $d$ to obtain the vertex-wise hue-shift vectors $\{d_k\}_{k=0}^{p}$, which can be formulated as:
\begin{subequations}
\label{eq:hue_shift}
\begin{align}
m_k &= \sum_{j=0}^{p} \hat{m}_j e^{-\tau |j-k|},
\label{eq:hue_shift-1} \\
d_k &= m_k \cdot d
\label{eq:hue_shift-2}
\end{align}
\end{subequations}
where $\{\hat{m}_k\}_{k=0}^{p}$ denote the learnable parameters with $\hat{m}_k \in \mathbb{R}$.

\subsection{Adaptive Cylindrical Color Transform and AdaLAB Color Space}
\label{sec:adacct}

Existing methods~\cite{LL.HVI-CIDNet} build a symmetric color space to address low-light color distortion. However, these reversible color transforms preserve identity throughout the transformation and \emph{do no more than reorganize the representation}, leaving the embedded color distortion alone. Consequently, the embedded color bias is carried into the enhancement process, further disrupting the balance between lightness recovery and color restoration, and finally manifests as color bias and abnormal saturation in the restored image.

To address this limitation, we propose AdaCCT together with AdaLAB color space. As shown in Fig.~\ref{fig:transform}, the forward AdaCCT maps the input image from RGB to AdaLAB through RGB-to-LAB conversion, hue-directional chromatic debiasing, and lightness-aware chroma scaling, where AdaLAB serves as the working space for subsequent backbone enhancement. After backbone enhancement in AdaLAB, the inverse AdaCCT reconstructs the enhanced RGB image by reverting chroma scaling, applying out-of-gamut lightness compensation, and converting the result from LAB to RGB.

\para{Hue-directional Chromatic Debiasing}
To suppress the embedded color bias before backbone enhancement, we incorporate a hue-directional chroma offset to move the color-shifted representation toward a corrected one.

Specifically, we first transform an input image $x\in\mathbb{R}^{3\times H\times W}$ to the LAB space, yielding a LAB map $x_\mathrm{l}\in\mathbb{R}^{3\times H\times W}$, which is then separated into a lightness map $l_\mathrm{l}\in\mathbb{R}^{1\times H\times W}$ and a chrominance map $u_{\mathrm{l}}\in\mathbb{R}^{2\times H\times W}$. After that, we apply a linear interpolation of the vertex-wise hue-shift vectors $\{d_k\}_{k=0}^{p}$ on the vertices $\{v\}^{p}$ according to the lightness map $l_\mathrm{l}$ to obtain lightness-aware hue-shift vectors $d_\mathrm{l}\in\mathbb{R}^{2\times H\times W}$. Then, we compute a similarity-aware weight $s_\mathrm{l}\in[\delta_1, \delta_2]^{1\times H\times W}$ between $u_\mathrm{l}$ and $d_\mathrm{l}$ to account for the boundary effect of color bias. Finally, the chrominance map $u_\mathrm{l}$ is shifted along the lightness-aware hue-shift vectors $d_\mathrm{l}$ with similarity weight $s_\mathrm{l}$. The above process is formulated as:
\begin{subequations}
\label{eq:forward_debias_image}
\begin{align}
d_\mathrm{l} &= \operatorname{Interp}\!\left(l_{\mathrm{l}};\{v_k\}_{k=0}^{p},\{d_k\}_{k=0}^{p}\right),
\label{eq:forward_debias_image_d} \\
s_\mathrm{l} &=\frac{\delta_2 - \delta_1}{2}\cdot\frac{u_\mathrm{l} \cdot d_\mathrm{l}}{\left \|u_\mathrm{l}\right \|_2 \left \|d_\mathrm{l}\right \|_2} + \frac{\delta_1 + \delta_2}{2},
\label{eq:forward_debias_image_s} \\
\tilde{u}_\mathrm{l} &= u_\mathrm{l} - s_\mathrm{l} \cdot d_\mathrm{l},
\label{eq:forward_debias_image_u}
\end{align}
\end{subequations}
where $\operatorname{Interp}(l_\mathrm{l};\{v\}^p;\{d\}^p)$ denotes the linear interpolation of the vertex values $\{d\}^{p}$ on the vertices $\{v\}^{p}$ according to the lightness map $l_\mathrm{l}$, $\delta_1$ and $\delta_2$ are fixed constants defining the range of the weighted similarity $s_\mathrm{l}$, \ie, $s_\mathrm{l}\in[\delta_1,\delta_2]$.

\para{Lightness-aware Chroma Scaling}
To provide a more structured representation for subsequent enhancement, we reorganize colors from different lightness planes into a more consistent chromatic range through AdaLAB color transform.

Specifically, we first obtain the lightness-aware chroma-scaling intensity $c_\mathrm{l}\in\mathbb{R}^{1\times H\times W}$ through a linear interpolation of the chroma-scaling intensities $\{c_k\}_{k=0}^{p}$ on the lightness sensitivity vertices $\{v_k\}_{k=0}^{p}$ according to the lightness map $l_\mathrm{l}$. We then multiply $\tilde{u}_\mathrm{l}$ by $c_\mathrm{l}$ to reorganize the chrominance representation and obtain the AdaLAB chrominance map $\hat{u}_\mathrm{l}\in\mathbb{R}^{2\times H\times W}$, which is formulated as:
\begin{subequations}
\label{eq:apply_scaling}
\begin{align}
c_\mathrm{l} &= \operatorname{Interp}\!\left(l_{\mathrm{l}};\{v_k\}_{k=0}^{p},\{c_k\}_{k=0}^{p}\right),
\label{eq:apply_scaling-1} \\
\hat{u}_\mathrm{l} &= c_\mathrm{l}\,\tilde{u}_\mathrm{l}.
\label{eq:apply_scaling-2}
\end{align}
\end{subequations}

We then merge the reorganized chrominance map $\hat{u}_\mathrm{l}$ and lightness map $l_\mathrm{l}$ into reorganized AdaLAB map $\hat{x}_\mathrm{l}\in\mathbb{R}^{3\times H\times W}$ for subsequent enhancement.

After backbone enhancement in the AdaLAB color space, we map the enhanced representation back to the base LAB space by reverting the lightness-aware chroma scaling. Formally, let the enhanced AdaLAB representation be $x_\mathrm{h}\in\mathbb{R}^{3\times H\times W}$, which is separated into an enhanced lightness map $\hat{l}_\mathrm{h}\in\mathbb{R}^{1\times H\times W}$ and an enhanced chrominance map $\hat{u}_\mathrm{h}\in\mathbb{R}^{2\times H\times W}$. We first obtain the lightness-aware chroma-scaling intensity $c_\mathrm{h}\in\mathbb{R}^{1\times H\times W}$ in the same way as $c_\mathrm{l}$, according to the enhanced lightness map $\hat{l}_\mathrm{h}$. We then divide $\hat{u}_\mathrm{h}$ by $c_\mathrm{h}+\epsilon$ to revert the chrominance representation back and obtain $\tilde{u}_\mathrm{h}\in\mathbb{R}^{2\times H\times W}$, which is formulated as:
\begin{subequations}
\label{eq:inverse_scaling_image}
\begin{align}
c_\mathrm{h} &= \operatorname{Interp}\!\left(\hat{l}_\mathrm{h};\{v_k\}_{k=0}^{p},\{c_k\}_{k=0}^{p}\right),
\label{eq:inverse_scaling_image-1} \\
\tilde{u}_\mathrm{h} &= \frac{\hat{u}_\mathrm{h}}{c_\mathrm{h}+\epsilon},
\label{eq:inverse_scaling_image-2}
\end{align}
\end{subequations}
where $\epsilon=1\times10^{-12}$ is a small constant for numerical stability.

\begin{figure}[t]
  \centering
  \includegraphics[width=\linewidth]{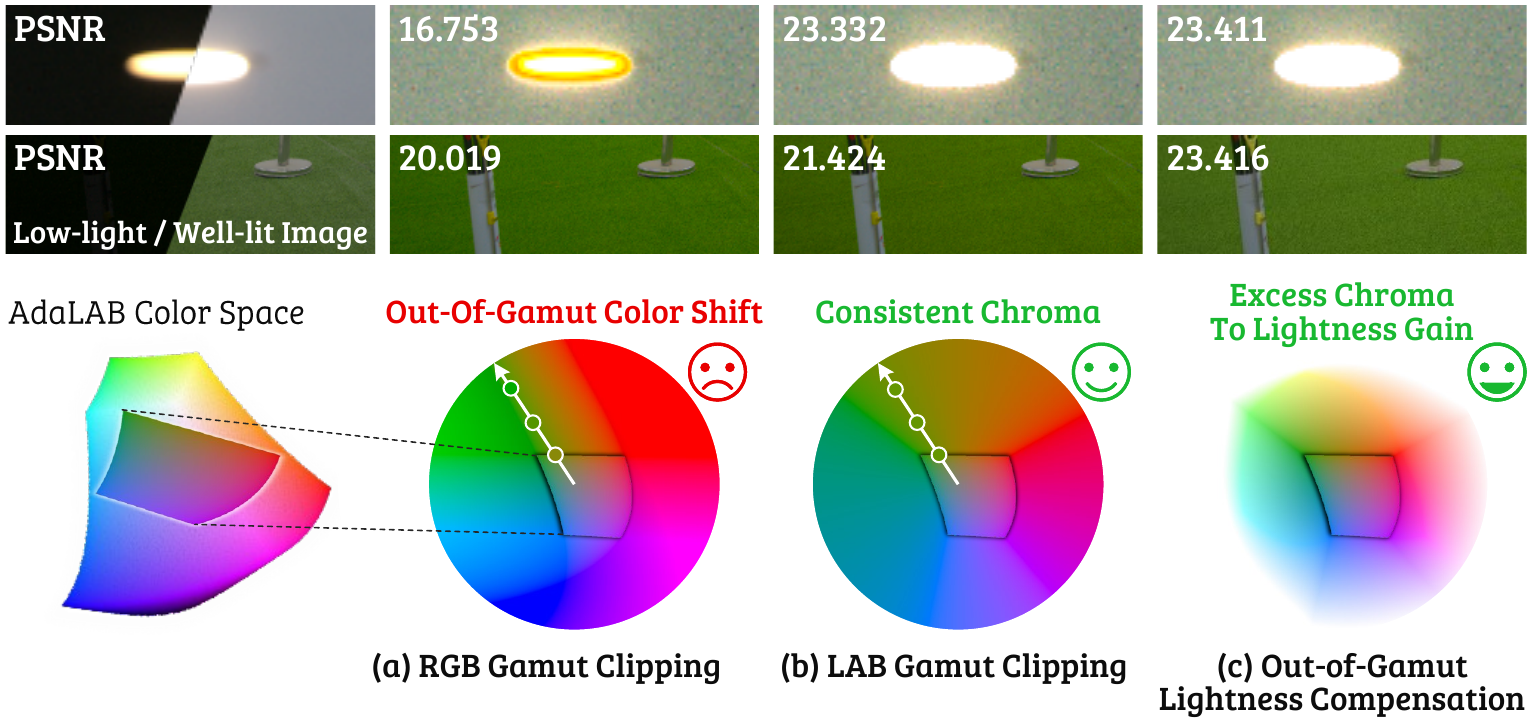}
  \caption{Comparison of Out-of-Gamut Handling Strategies. }
  \Description{
  The figure compares RGB gamut clipping, LAB gamut clipping, and out-of-gamut lightness compensation on two low-light examples. RGB gamut clipping introduces visible color shifts, LAB gamut clipping preserves a more consistent chroma direction, and out-of-gamut lightness compensation converts excess chroma into lightness gain while retaining valid colors.
  }
  \label{fig:gamut}
  \vspace{-2mm}
\end{figure}

\para{Out-of-gamut Lightness Compensation}
After reverting the chroma scaling, the restored LAB representation is mapped back to RGB. A key issue at this stage is that the restored color may fall outside the valid RGB gamut. Direct gamut clipping would truncate excessive chroma at the boundary and easily create visible color bias artifacts in highly saturated regions, as shown in Fig.~\ref{fig:gamut}.

To address this issue, we introduce out-of-gamut lightness compensation (OOGLC). In detail, we first clip the original chrominance map $\tilde{u}_\mathrm{h}$ to the valid gamut according to $\hat{l}_\mathrm{h}$, which yields a clipped chrominance map $u_\mathrm{c}\in\mathbb{R}^{2\times H\times W}$. The difference between $u_\mathrm{c}$ and $\tilde{u}_\mathrm{h}$ reflects the amount of chroma that cannot be represented within the valid gamut. Rather than discarding this out-of-gamut chroma, we convert it into lightness compensation to obtain a lightness-saturation balanced reconstruction. After that, $u_\mathrm{c}$ is clipped to the valid gamut according to $l_\mathrm{h}$ to ensure color validity. The above process can be mathematically expressed as:
\begin{subequations}
\label{eq:OOGLC}
\begin{align}
u_\mathrm{c} &= \operatorname{GamutClipping}(\tilde{u}_\mathrm{h}, \hat{l}_\mathrm{h}),
\label{eq:OOGLC-1} \\
l_\mathrm{h} &= \hat{l}_\mathrm{h} + \gamma \|u_\mathrm{c} - \tilde{u}_\mathrm{h}\|_{2},
\label{eq:OOGLC-2} \\
u_\mathrm{h} &= \operatorname{GamutClipping}(u_\mathrm{c}, l_\mathrm{h}),
\label{eq:OOGLC-3}
\end{align}
\end{subequations}
where $\operatorname{GamutClipping}(\cdot,\cdot)$ clips a color along the saturation direction to the valid gamut, and $\gamma=1.0$ denotes the compensation ratio. Following previous work~\cite{LL.HVI-CIDNet}, we further introduce two customizing parameters, $\alpha_c$ and $\alpha_l$, to adjust image saturation and brightness, respectively:
\begin{equation}
\hat{y}_\mathrm{h} = [\alpha_\mathrm{l} l_\mathrm{h}; \alpha_\mathrm{c} u_\mathrm{h}],
\end{equation}
where $[\cdot;\cdot]$ denotes channel-wise concatenation. The final valid LAB map $\hat{y}_\mathrm{h}$ is then converted back to the RGB space through the inverse color space mapping, obtaining the enhanced image $\hat{y}$.

\subsection{Training Objectives}
To constrain restoration in both the AdaLAB and RGB spaces, we supervise the enhanced AdaLAB map $\hat{y}_\mathrm{AdaLAB}$ and the restored RGB image $\hat{y}$ with their corresponding targets. Specifically, $y_\mathrm{AdaLAB}$ is obtained by mapping the ground-truth image $y$ to the base LAB space and applying the same lightness-aware chroma scaling predicted from the input low-light image $x$, without the hue-directional chroma offset. The total loss is defined as:
\begin{equation}
    L_{\mathrm{total}}=\lambda \cdot L(\hat{y}_\mathrm{AdaLAB}, y_\mathrm{AdaLAB}) + L(\hat{y}, y),
\end{equation}
where $\lambda$ balances the two losses, and $L(\cdot,\cdot)$ is the reconstruction loss of the baseline network.

\input{tables/main-lol_results}

\begin{figure*}[h]
  \centering
  \includegraphics[width=\linewidth]{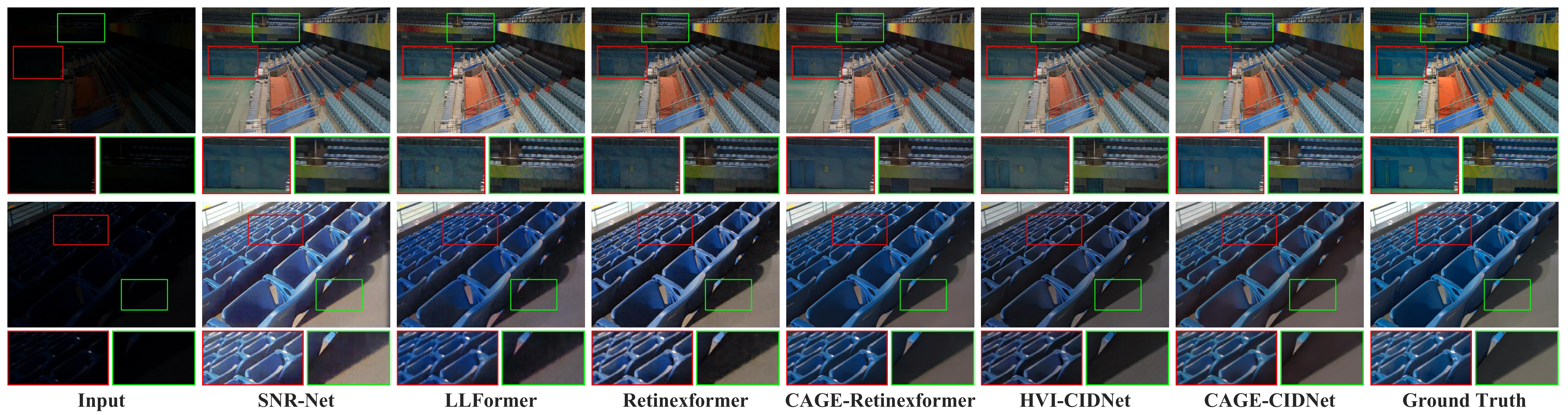}
  \caption{Qualitative comparison of representative methods and baseline methods with and without our proposed CAGE on LOLv1 dataset (first row) and LOLv2-real dataset (second row). Our CAGE empowers baseline methods to produce images with less color distortion, \eg, less saturation abnormality in the first row, less color bias in the second row.}
  \Description{
  The figure compares input images, representative enhancement methods, the corresponding backbones with CAGE, and ground-truth images. The first row shows that CAGE reduces local saturation abnormality, while the second row shows that CAGE suppresses the residual color bias and produces colors closer to the ground truth.
  }
  \label{fig:lolv2-vis}
\end{figure*}

\section{Experiments}

\subsection{Experimental Setup}

\para{Implementation Details}
We run all experiments on NVIDIA A40 GPUs. The backbone networks used in our method are retrained for fair comparison based on their released code. For the other comparison methods, we directly use the published results when available, and retrain them with their official implementations on the datasets where their results are not provided.

\para{Baselines}
To comprehensively evaluate the effectiveness of CAGE, we compare with representative low-light image enhancement methods~\cite{LL.RetinexNet.Dataset.LOLv1, LL.KinD, LL.DRBN, IR.MIRNet, LL.Zero-DCE, LL.EnlightenGAN, LL.LLFlow, LL.SNRNet, LL.BreaD, LL.FourLLIE, LL.LLFormer, LL.QuadPrior, IR.URWKV, LL.CWNet, LL.Retinexformer, LL.DarkIR, LL.HVI-CIDNet}. To further verify the compatibility of CAGE with different enhancement backbones, we integrate it into three representative methods, including Retinexformer~\cite{LL.Retinexformer} for Retinex-based enhancement, DarkIR~\cite{LL.DarkIR} for RGB-space enhancement, and HVI-CIDNet~\cite{LL.HVI-CIDNet} for HSV-inspired color-space enhancement. More details of the integration for each backbone are provided in Appendix~\ref{sec:supp:integration}.

\para{Datasets and Metrics}
We evaluate CAGE on six paired low-light benchmarks, including LOLv1~\cite{LL.RetinexNet.Dataset.LOLv1}, LOLv2-real~\cite{Dataset.LOLv2}, LOLv2-synthetic~\cite{Dataset.LOLv2}, SDSD-indoor~\cite{Dataset.SDSD}, SDSD-outdoor~\cite{Dataset.SDSD}, and SID~\cite{Dataset.LL.SID}.
Following common practice of~\cite{LL.HVI-CIDNet, LL.Retinexformer}, we use PSNR~\cite{PSNR}, SSIM~\cite{SSIM}, and LPIPS~\cite{LPIPS} to evaluate the restored images on paired datasets.
Additional experimental results and setups are provided in Appendix~\ref{sec:supp:exp}.

\subsection{Quantitative Results}

To evaluate the effectiveness of CAGE for color-faithful low-light image enhancement, we integrate it into representative enhancement backbones and compare the resulting models with both their original baselines and other competitive methods.

\para{Standard Benchmarks} LOLv1 and LOLv2 are standard paired benchmarks for low-light image enhancement. As shown in Table~\ref{tab:main_results_lol}, CAGE consistently improves all backbones across LOLv1 and LOLv2 datasets. Compared with the original backbones, CAGE achieves an average PSNR gain of 0.98 dB / 1.23 dB / 0.81 dB on LOLv1, LOLv2-real, and LOLv2-synthetic, respectively.
This improvement is achieved with only \textbf{0.07M} additional parameters (about 3.5\% of HVI-CIDNet) and \textbf{less than 0.01G} FLOPs (less than 0.1\% of HVI-CIDNet), indicating that the performance gain mainly comes from improved color modeling rather than increased model capacity. Compared with RGB-based method (\ie, DarkIR), Retinex-based method (\ie, Retinexformer), and even HSV-inspired color-space enhancement (\ie, HVI-CIDNet), CAGE brings substantial improvements on all evaluated datasets. This demonstrates that explicitly reorganizing chromatic distribution provides more stable color restoration under strong enhancement.

\para{Challenging Scenarios} SDSD and SID pose a more difficult test for low-light image enhancement due to large illumination variation and severe low-light degradation with strong color bias. As reported in Table~\ref{tab:main_results_sdsd_sid}, CAGE achieves consistent improvement on both SDSD and SID datasets. Compared with strong baselines such as DarkIR, Retinexformer, and HVI-CIDNet, CAGE improves PSNR by 1.12 dB on SDSD-indoor, 0.73 dB on SDSD-outdoor, and 0.56 dB on SID.
This indicates that the proposed cylindrical transform effectively aligns chromatic distributions across lightness levels, thereby mitigating instability from spatially varying illumination. Concurrently, suppressing embedded color bias is shown to be essential for scenes with extreme lighting disparity.

\subsection{Qualitative Results}
To further verify the visual benefit of CAGE, we compare the restored images on real-world low-light scenes. The visual comparisons on LOLv1 and LOLv2-real are shown in Fig.~\ref{fig:lolv2-vis}. Compared with the original backbone outputs, the images enhanced by CAGE show less global color bias, more stable local saturation, and more natural chromatic transitions in dark regions and detail-rich regions. In scenes with visible color cast, CAGE variants (\ie, CAGE-Retinexformer, CAGE-DarkIR, and CAGE-CIDNet) restore more faithful neutral regions and suppress the residual tint left by the baseline backbone. In scenes with strong enhancement, CAGE also reduces washed-out colors and oversaturated regions, which leads to a more balanced visual result. More visual comparisons are provided in Appendix~\ref{sec:supp:benchmark}.

\input{tables/main-sdsd_results}

\subsection{Ablation Study}

We conduct several ablation experiments on real-world dataset (\ie, LOLv2-real) to evaluate the contribution of each component in CAGE and analyze their impact on color-faithful low-light image enhancement.

\para{Cylindrical Adaptive Parameters}
Table~\ref{tab:abl:cylindrical} shows that the full parameter set (Table~\ref{tab:abl:cylindrical} \circnum{4}) achieves the best results on all metrics. Removing hue-shift parameters (Table~\ref{tab:abl:cylindrical} \circnum{1}) causes the largest drop in PSNR and SSIM, reducing PSNR by 0.4862 dB and SSIM by 0.0107, while increasing LPIPS by 0.0003. Removing chroma-scaling intensity (Table~\ref{tab:abl:cylindrical} \circnum{2}) leads to a smaller PSNR drop of 0.2118 dB and SSIM by 0.0012, but causes the largest LPIPS increase of 0.0069. Removing lightness sensitivity (Table~\ref{tab:abl:cylindrical} \circnum{3}) keeps PSNR close to that of the full model, but still leads to inferior SSIM and LPIPS. These results show that hue-shift correction is the key part for suppressing low-light color bias, while chroma scaling and lightness sensitivity further improve structural consistency and perceptual quality.

\begin{figure}[t]
  \centering
  \includegraphics[width=\linewidth]{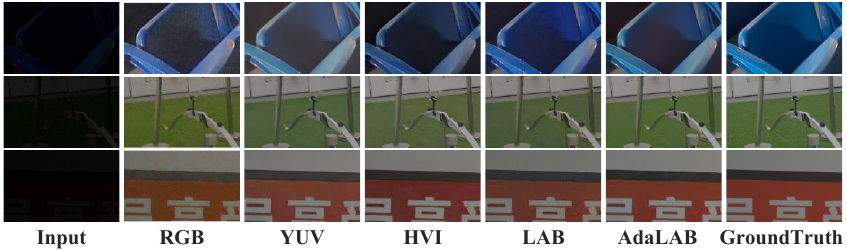}
  \vspace{-6mm}
  \caption{Visual comparison of different color spaces for LLIE. }
  \Description{
  Three low-light examples are enhanced in the RGB, YUV, HVI, LAB, and AdaLAB color spaces and compared with their ground-truth images. AdaLAB produces more consistent brightness and chromatic distributions, with less color bias and saturation abnormality than the other color spaces.
  }
  \label{fig:abl:color-space}
\end{figure}

\para{AdaLAB Color Space}
Our proposed AdaLAB color space provides a more coherent basis than YUV and the HSV-inspired (\ie, HVI) space for modeling low-light-induced color shift. As shown in Table~\ref{tab:abl:color_space}, AdaLAB (Table~\ref{tab:abl:color_space} \circnum{5}) achieves the best results among all compared spaces, reaching 24.2355 PSNR, 0.8746 SSIM, and 0.0977 LPIPS. Among the baseline spaces, LAB (Table~\ref{tab:abl:color_space} \circnum{4}) gives the strongest performance, surpassing RGB (Table~\ref{tab:abl:color_space} \circnum{1}) and YUV (Table~\ref{tab:abl:color_space} \circnum{2}) by clear margins. Compared with HVI (Table~\ref{tab:abl:color_space} \circnum{3}), LAB improves PSNR by 0.0807 dB and SSIM by 0.0006, while reducing LPIPS by 0.0079. Built on LAB, AdaLAB further improves PSNR by 0.2184 dB, SSIM by 0.0028, and reduces LPIPS by 0.0072. These results indicate that LAB provides a better basis for color debiasing by more effectively decoupling lightness and chrominance, while AdaLAB further improves chromatic restoration through lightness-aware chroma alignment. As shown in Fig.~\ref{fig:abl:color-space}, AdaCCT in AdaLAB yields more faithful colors.

\para{Gamut Harmonization}
As shown in Table~\ref{tab:abl:gamut}, replacing RGB gamut clipping (Table~\ref{tab:abl:gamut} \circnum{1}) with LAB gamut clipping (Table~\ref{tab:abl:gamut} \circnum{2}) significantly improves the results, and out-of-gamut lightness compensation (Table~\ref{tab:abl:gamut} \circnum{3}) further improves them to 24.2355 PSNR, 0.8746 SSIM, and 0.0977 LPIPS.
These results show that simple gamut clipping fails to recover chroma adequately, whereas out-of-gamut lightness compensation achieves more faithful chroma recovery.

\input{tables/main-ablation}

\section{Conclusion}

In this paper, we revisit low-light image enhancement from the perspective of embedded chromatic disturbance, and show that lightness-chrominance decoupling alone is insufficient to eliminate the color bias inherited from low-light inputs. To address this issue, we propose CAGE, a cylindrical color correction framework built upon the proposed AdaLAB color space and the adaptive transform AdaCCT. The forward transform suppresses the embedded color bias before backbone enhancement by reorganizing the chromatic distribution, while the inverse transform stabilizes color reconstruction through out-of-gamut lightness compensation.
Extensive experiments on multiple benchmarks demonstrate that CAGE consistently improves color fidelity, alleviates both global color bias and local saturation abnormality, and enhances overall visual quality across different low-light enhancement backbones. These results verify that explicitly modeling and correcting chromatic disturbance in a decoupled cylindrical representation provides a more reliable solution for color-faithful low-light image enhancement.

\begin{acks}
This work was supported in part by the National Natural Science Foundation of China under Grant 62471142, in part by the Industry-Academy Cooperation Project under Grant 2024H6006, in part by the 2025 Special Fund Project for Promoting High-Quality Development of Marine and Fishery Industries in Fujian Province under Grant FJHYF-L-2025-07-004, and in part by the Fujian Provincial Science and Technology Program Project under Grant 2026Y0090.
\end{acks}

\bibliographystyle{ACM-Reference-Format}
\balance
\bibliography{sample-base}

\clearpage
\appendix

This appendix presents two central aspects of \FRAMEWORK: \emph{i)} the design rationale of the proposed adaptive cylindrical color transform (AdaCCT) in Sec.~\ref{sec:supp:design}, and further explains the adaptive interpolation on the lightness axis in Sec.~\ref{sec:supp:interp} and the integration into different low-light enhancement frameworks in Sec.~\ref{sec:supp:integration}. \emph{ii)} the framework generalization of the proposed color modeling with training details in Sec.~\ref{sec:supp:exp_setup}, including experimental environment and hyperparameter configuration, dataset details in Sec.~\ref{sec:supp:datasets}, more benchmarking results in Sec.~\ref{sec:supp:benchmark}, where \textbf{our proposed \FRAMEWORK\ further demonstrates strong generalization capability and clear performance gains on no-reference low-light benchmarks}, and more ablation results in Sec.~\ref{sec:supp:abl}.

\section{Design Details and Visual Analyses}
\label{sec:supp:design}

\begin{figure*}[t]
  \centering
  \includegraphics[width=0.9\textwidth]{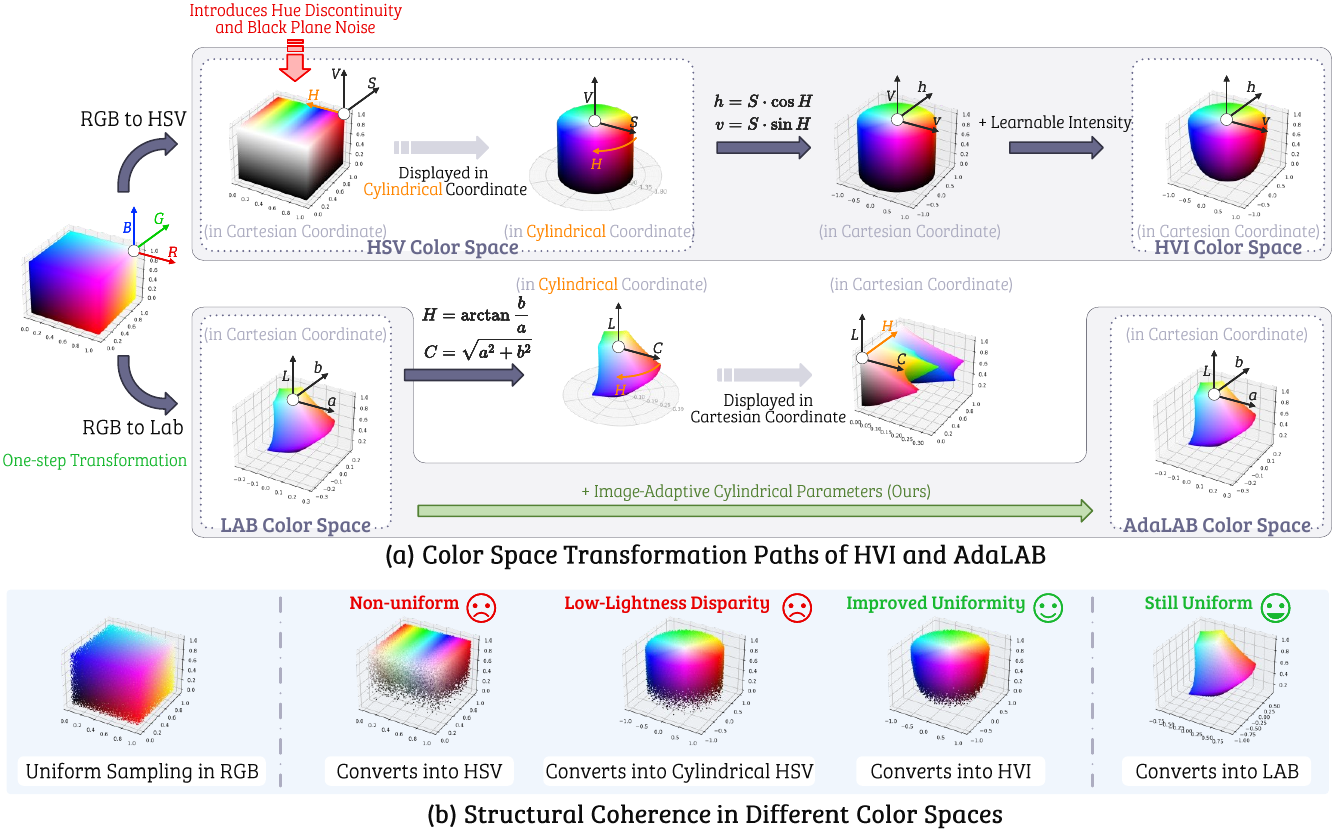}
  \caption{LAB serves as a more coherent basis for color-faithful low-light image enhancement. \emph{(a)} HVI follows the RGB-to-HSV route and inherits the structural weakness of HSV, while AdaLAB starts from the one-step RGB-to-LAB conversion and further introduces image-adaptive cylindrical parameters. \emph{(b)} Under uniform RGB sampling, HSV becomes non-uniform and cylindrical HSV shows clear low-lightness disparity, whereas LAB preserves a more coherent organization. }
  \Description{}
  \label{fig:supp:color_space_overview}
  \vspace{4mm}
\end{figure*}

Rather than preserving identity throughout a reversible transformation, \TRANSFORM\ is built to correct the embedded chromatic disturbance in low-light images. In the forward transform, the input image is mapped from RGB to AdaLAB through RGB-to-LAB conversion, hue-directional chromatic debiasing, and lightness-aware chroma scaling, so that the embedded color bias is reduced before backbone enhancement. In the inverse transform, the enhanced result is mapped back to RGB by reverting chroma scaling, applying out-of-gamut lightness compensation, and converting the result from LAB to RGB. The same asymmetry is kept in the \LAB-space supervision, where the target $y_{\mathrm{AdaLAB}}$ shares the lightness-aware chroma scaling predicted from the input low-light image $x$, but does not include the hue-directional chroma offset.

To further unpack the design of \TRANSFORM, we first revisit the color-space path underlying \LAB. As shown in Fig.~\ref{fig:supp:color_space_overview}, the RGB-to-HSV route can separate brightness from chromatic content, but the conversion also brings hue discontinuity and black-plane noise, and the cylindrical HSV form further exhibits clear low-lightness disparity. By comparison, as reflected in Fig.~\ref{fig:supp:color_space_overview}\emph{(b)}, the one-step RGB-to-LAB conversion preserves a more uniform and structurally coherent organization of RGB-sampled points, whereas HSV becomes non-uniform and introduces low-lightness disparity, making LAB a cleaner basis for building AdaLAB and introducing the image-adaptive cylindrical parameters. The remaining parts of this section complement this overview from two aspects: the interpolation from lightness vertices to pixel-wise values, and the integration of AdaCCT into different enhancement frameworks.

\subsection{Lightness-aware Linear Interpolation}
\label{sec:supp:interp}
The lightness-aware linear interpolation can be described as \textit{the linear interpolation of the vertex values on the vertices according to the lightness map}. 

\begin{figure}[t]
  \centering
  \includegraphics[width=\linewidth]{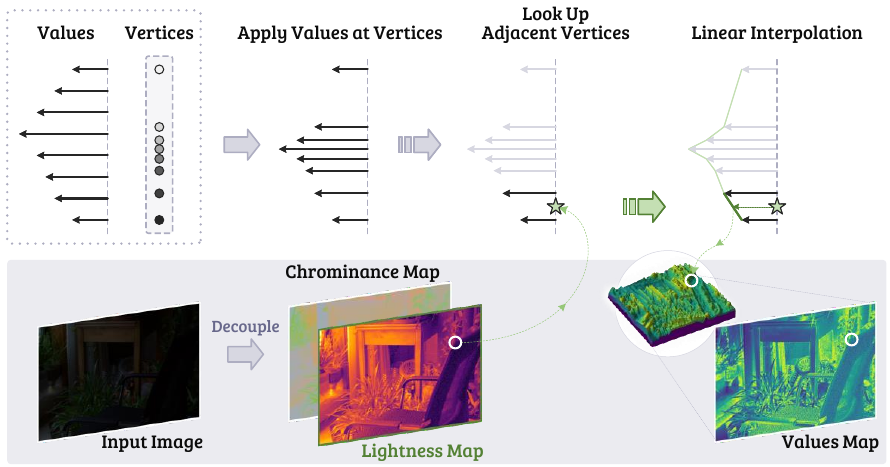}
  \caption{Detailed pipeline of the lightness-aware interpolation operation. Given the lightness map, we first place the values at the lightness sensitivity vertices along the lightness axis, then determine the two neighboring vertices for each pixel according to the lightness map, and finally apply linear interpolation to obtain the lightness-aware value map. 
  }
  \Description{}
  \label{fig:supp:interpolation}
\end{figure}

As shown in Fig.~\ref{fig:supp:interpolation}, given a lightness map $l \in \mathbb{R}^{1 \times H \times W}$, lightness sensitivity vertices $\{v_k\}_{k=0}^{p}$, and vertex values $\{r_k\}_{k=0}^{p}$, the interpolation operator produces a dense lightness-aware map $a \in \mathbb{R}^{C \times H \times W}$, which can be formulated as:
\begin{equation}
a = \mathrm{Interp}(l; \{v_k\}_{k=0}^{p}, \{r_k\}_{k=0}^{p}),
\end{equation}
where $C=2$ for hue-shift vectors and $C=1$ for chroma-scaling intensity. For each pixel location $(i,j)$, let $l_{ij}$ denote the lightness value at that pixel. We first locate the two adjacent lightness sensitivity vertices surrounding $l_{ij}$ along the lightness axis. The index of the lower vertex is denoted by $q_{ij}$, where $q_{ij} = \max \{ k \mid v_k \le l_{ij},\ 0 \le k \le p-1 \}$. Since $v_0 = 0$ and $v_p = 1$, each lightness value lies in the interval $[v_{q_{ij}}, v_{q_{ij}+1}]$ defined by these two adjacent vertices. We then compute the relative position of $l_{ij}$ within this interval using the interpolation weight $t_{ij} = \frac{l_{ij} - v_{q_{ij}}}{v_{q_{ij}+1} - v_{q_{ij}}}$. Based on this relative position, the pixel-wise value is linearly interpolated from the two neighboring vertex values as $a_{ij} = (1 - t_{ij}) r_{q_{ij}} + t_{ij} r_{q_{ij}+1}$. When $r_k$ is vector-valued, the same interpolation is applied channel-wise.

With this formulation, the same operator is shared by different parts of \TRANSFORM. In the forward transform, the lightness-aware hue-shift vectors $d_l$ and the lightness-aware chroma-scaling intensity $c_l$ are obtained as
\begin{subequations}
\begin{align}
d_l &= \mathrm{Interp}(l_l; \{v_k\}_{k=0}^{p}, \{d_k\}_{k=0}^{p}), \\
c_l &= \mathrm{Interp}(l_l; \{v_k\}_{k=0}^{p}, \{c_k\}_{k=0}^{p}).
\end{align}
\end{subequations}
In the inverse transform, the same interpolation is applied to the enhanced lightness map $\hat{l}_h$ to obtain
\begin{equation}
c_h = \mathrm{Interp}(\hat{l}_h; \{v_k\}_{k=0}^{p}, \{c_k\}_{k=0}^{p}).
\end{equation}

\subsection{Integration Strategies for Difference LLIE Frameworks}
\label{sec:supp:integration}

\begin{figure}[t]
  \centering
  \includegraphics[width=\linewidth]{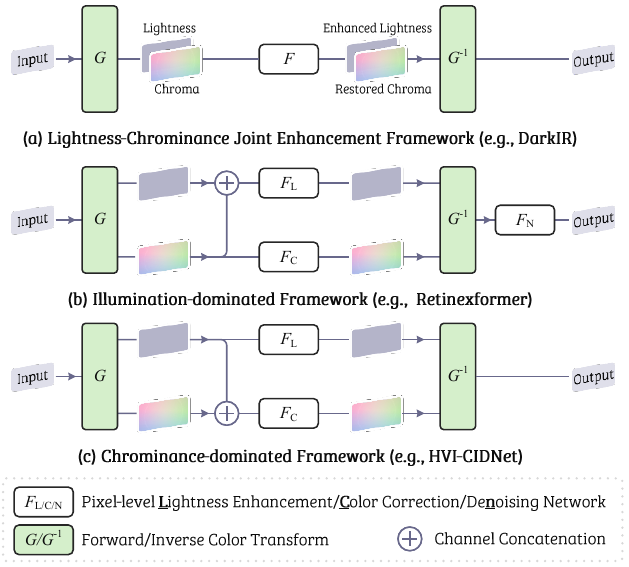}
  \caption{Integration strategies of \FRAMEWORK\ in three representative low-light enhancement paradigms: \emph{(a)} Lightness-chrominance joint enhancement frameworks, where \FRAMEWORK\ replaces RGB processing with the proposed forward and inverse transforms; \emph{(b)} Illumination-dominated frameworks, where \FRAMEWORK\ replaces the original decomposition and recomposition pipeline in the transformed space; and \emph{(c)} Chrominance-donimated enhancement frameworks, where \FRAMEWORK\ is placed around the decoupled lightness and chroma branches before reconstruction.
  }
  \Description{}
  \label{fig:integration_strategy}
\end{figure}

Fig.~\ref{fig:integration_strategy} illustrates how \FRAMEWORK\ is integrated into three representative LLIE framework paradigms, including the lightness-chrominance joint enhancement framework, the illumination-dominated Retinex-based enhancement framework, and the chrominance-dominated HSV-inspired color-space enhancement framework. Rather than changing the main enhancement branches, \FRAMEWORK\ adapts the original pipeline by introducing the forward transform $G$ and inverse transform $G^{-1}$ around the enhancement process. In particular, the input image is first mapped into the transformed space for lightness and chromatic processing, and the enhanced transformed output is then mapped back to RGB, with the original joint enhancement, branch interaction, or post-reconstruction processing preserved according to the corresponding framework design.

\para{Lightness-Chrominance Joint Enhancement Framework}
RGB-space enhancement~\cite{LL.DarkIR} designs provide a representative integration setting for this paradigm. The original RGB enhancement pipeline is replaced by the transformed-space pipeline of \FRAMEWORK. The input image $x$ is first mapped into the transformed space by the forward transform, the backbone network then enhances the transformed representation jointly, and the enhanced result is finally mapped back to RGB by the inverse transform:
\begin{equation}
\hat{y} = G^{-1}\!\left(F\!\left(G(x)\right)\right),
\end{equation}
where $x_\mathrm{l}=G(x)$ denotes the transformed input representation, which consists of the input lightness map $l$ and chromatic map $u$, and $x_\mathrm{h}=F(x_\mathrm{l})$ denotes the enhanced transformed representation, which consists of the enhanced lightness map $\hat{l}$ and chromatic map $\hat{u}$. The final RGB image is reconstructed as $\hat{y}=G^{-1}(x_\mathrm{h})$.

However, RGB images entangle lightness and chromatic information, and the original pipeline does not account for explicit lightness-chrominance decoupling. Directly feeding decoupled lightness and chromatic components into such backbones can instead be counterproductive. To better match this processing pattern, we further introduce a pseudo Retinex-based decoupling and recomposition around the transformed-space backbone. Specifically, the transformed chromatic map is first normalized by the lightness map to obtain a pseudo reflectance-related component $u_\mathrm{l} = u \oslash l$. The enhancement backbone then takes $l$ and $u_\mathrm{l}$ as input and produces the enhanced lightness map $\hat{l}$ and the enhanced pseudo reflectance-related component $u_\mathrm{h}$. Before the inverse transform, the transformed chromatic map is recovered by $\hat{u} = u_\mathrm{h} \odot \hat{l}$. The whole process can be formulated as:
\begin{subequations}
\begin{align}
(l, u) &= G(x),\\
(\hat{l}, \hat{u}) &= F([l; u \oslash l]),\\
\hat{y} &= G^{-1}([\hat{l}; \hat{u} \odot \hat{l}]),
\end{align}
\end{subequations}
where $u_\mathrm{l}$ and $u_\mathrm{h}$ denote the pseudo reflectance-related component before and after enhancement, respectively, $\oslash$ denotes element-wise division between the chromatic map and the lightness map, and $\odot$ denotes element-wise multiplication for chromatic map recovery, $[\cdot;\cdot]$ denotes channel-wise concatenation.

\para{Illumination-dominated Framework} 
A representative integration setting for this paradigm is given by Retinex-based enhancement frameworks~\cite{LL.Retinexformer}. 
In this paradigm, the enhancement pipeline is organized as a decomposition-and-recomposition process centered on lightness recovery. Specifically, the input image is first decomposed into a lightness-related component $l$ and a reflectance-related component $u$ via $x = l \odot u$. The lightness branch then focuses on brightness recovery, while the reflectance branch preserves reflectance consistency. After branch-wise enhancement, the enhanced components are recomposed into an RGB image via $\tilde{y} = \hat{l} \odot \hat{u}$, and a denoising network can be further appended after recomposition in frameworks with such a design.

Our proposed \FRAMEWORK\ replaces the original decomposition with the forward transform and the original recomposition with the inverse transform, and keeping the post-reconstruction denoising network unchanged. 
Specifically, the input image is first mapped by $G$ into the transformed space to obtain the lightness map and chromatic map. The transformed lightness map is then enhanced by the lightness branch $F_\mathrm{L}$ with the transformed chromatic map as input, while the transformed chromatic map is enhanced by the chromatic branch $F_\mathrm{C}$. The enhanced lightness map and chromatic map are then mapped back to RGB by the inverse transform $G^{-1}$. The whole process can be mathematically expressed as:
\begin{subequations}
\begin{align}
(l, u) &= G(x),\\
(\hat{l}, \hat{u}) &= \bigl(F_\mathrm{L}([l; u]),\, F_\mathrm{C}(u)\bigr),\\
\tilde{y} &= G^{-1}([\hat{l}; \hat{u}]),
\end{align}
\end{subequations}
where $(l, u)$ denote the transformed input with the lightness map and chromatic map, $(\hat{l}, \hat{u})$ denote the enhanced transformed output, and $\tilde{y}$ denotes the reconstructed RGB image before the optional denoising network, $[\cdot;\cdot]$ denotes channel-wise concatenation.
For frameworks with a denoising network after reconstruction~\cite{LL.Retinexformer}, the reconstructed RGB image is further refined by $F_\mathrm{N}$ to obtain the final enhanced output:
\begin{equation}
\hat{y} = F_\mathrm{N}(\tilde{y}),
\end{equation}
where $\hat{y}$ denotes the final enhanced output.

\para{Chrominance-dominated Framework} 
A representative integration setting for this paradigm is given by HSV-inspired color-space enhancement frameworks~\cite{LL.HVI-CIDNet}. In this paradigm, \FRAMEWORK\ is inserted around the decoupled lightness and chromatic branches before reconstruction, preserving the original decoupled enhancement path while replacing only the outer transform and RGB reconstruction process. Specifically, the input image is first mapped by $G$ into the transformed space to obtain the lightness map and chromatic map. The lightness map is then enhanced by the lightness branch $F_\mathrm{L}$, while the chromatic map is enhanced by the chromatic branch $F_\mathrm{C}$ with the lightness map as additional input. The enhanced lightness map and chromatic map are finally mapped back to RGB by the inverse transform. The entire process can be formulated as:
\begin{subequations}
\begin{align}
(l, u) &= G(x),\\
(\hat{l}, \hat{u}) &= \bigl(F_\mathrm{L}(l),\, F_\mathrm{C}([u; l])\bigr),\\
\hat{y} &= G^{-1}([\hat{l}; \hat{u}]),
\end{align}
\end{subequations}
where $(l, u)$ denote the transformed input with the lightness map and chromatic map, $(\hat{l}, \hat{u})$ denote the enhanced transformed output, and $\hat{y}$ denotes the reconstructed RGB image, $[\cdot;\cdot]$ denotes channel-wise concatenation.

These three integration strategies provide a unified way to integrate \FRAMEWORK\ into various low-light image enhancement frameworks, while keeping their original backbone structures unchanged.

\section{More Experiments}
\label{sec:supp:exp}

\subsection{Training Details}
\label{sec:supp:exp_setup}

\para{Experimental Environment}
The experiments were run on a server equipped with an Intel Xeon Gold 6326 CPU (16 cores @ 2.90GHz), NVIDIA A40 GPU (48GB VRAM), and 251GB system memory, running Ubuntu 20.04.6 LTS. The software stack included CUDA 11.6, Python 3.8.20, GCC 9.4.0, and PyTorch 1.13.0. Full dependency specifications is provided in the open-source release of our code.

\para{Hyperparameter Configuration}
Unless otherwise specified, we set the number of lightness intervals to $p=32$, the response range of the similarity-aware weight to $\delta_1=0.2$ and $\delta_2=1.0$, the lightness compensation ratio in out-of-gamut lightness compensation to $\gamma=1.0$, and the loss balance between the \LAB\ reconstruction loss and the RGB reconstruction loss to $\lambda=1.0$. For the final linear adjustment before LAB-to-RGB conversion, we set $\alpha_\mathrm{l}=1.0$ and $\alpha_\mathrm{c}=1.0$ by default. Following previous work~\cite{LL.HVI-CIDNet}, we use dataset-specific brightness and saturation adjustment on LOLv1 and LOLv2-Real. For LOLv1, we set $\alpha_\mathrm{l}=1.3$ and $\alpha_\mathrm{c}=1.0$. For LOLv2-Real, we set $\alpha_\mathrm{l}=1.1$ and $\alpha_\mathrm{c}=0.8$.

\para{GT Mean Evaluation}
Since LOLv1 contains only 15 testing pairs, direct metric evaluation can be sensitive to global brightness fluctuation and may obscure the comparison of color restoration and structural recovery. Following common practice in low-light image enhancement~\cite{LL.HVI-CIDNet, LL.HVI-CIDNet+, LL.GLARE, LL.LLFlow}, we use GT mean evaluation on LOLv1 during testing. For each predicted image, we first convert both the prediction and the ground-truth image into grayscale to compute their mean luminance, and then linearly rescale the prediction to match the mean luminance of the ground truth before metric evaluation. This process can be formulated as follows:
\begin{equation}
\hat{y}_{\mathrm{GTmean}} = \hat{y} \cdot \frac{\mu(y)}{\mu(\hat{y})},
\end{equation}
where $\hat{y}\in\mathbb{R}^{3\times H\times W}$ denotes the enhanced image, $y\in\mathbb{R}^{3\times H\times W}$ denotes the ground-truth image, and $\mu(\cdot)$ denotes the mean grayscale operation computed by averaging all pixel values after grayscale conversion. This operation reduces unnecessary luminance variation in the final comparison and makes PSNR, SSIM, and LPIPS on LOLv1 more focused on restoration quality beyond global brightness mismatch.

\begin{table}[t]
  \centering
  \caption{Summary of the datasets used in the experiments.}
  \label{tab:supp:datasets}
  \setlength{\tabcolsep}{8pt}
  \renewcommand{\arraystretch}{1.08}
  \resizebox{\linewidth}{!}{
  \begin{tabular}{c!{\vrule}l!{\vrule}cc!{\vrule}l}
    \toprule
    \rowcolor{gray}
    \textbf{Dataset Group} & \textbf{Dataset / Subset} & \textbf{\#Train} & \textbf{\#Test} & \textbf{Resolution} $(H \times W)$ \\
    \midrule
    \multirow{3}{*}{Paired LOL}
      & LOLv1           & 485  & 15  & $400 \times 600$ \\
      & LOLv2-Real      & 689  & 100 & $400 \times 600$ \\
      & LOLv2-Synthetic & 900  & 100 & $384 \times 384$ \\
    \midrule
    \multirow{2}{*}{Paired SDSD}
      & SDSD-indoor     & 1655 & 308 & $512 \times 960$ \\
      & SDSD-outdoor    & 2650 & 500 & $512 \times 960$ \\
    \midrule
    Paired SID
      & Sony subset     & 2099 & 598 & $1424 \times 2128$ \\
    \midrule
    \multirow{7}{*}{Unpaired}
      & DICM            & 0    & 69  & Various \\
      & LIME            & 0    & 10  & Various \\
      & MEF             & 0    & 17  & Various \\
      & NPE             & 0    & 8   & Various \\
      & VV              & 0    & 24  & Various \\
    \bottomrule
  \end{tabular}}
\end{table}

\input{tables/supp-unpaired_results}

\subsection{Datasets}
\label{sec:supp:datasets}

Detailed information on all datasets is summarized in Table~\ref{tab:supp:datasets}. In this appendix, we further include more visual comparisons on LOLv1~\cite{LL.RetinexNet.Dataset.LOLv1}, LOLv2-Real~\cite{Dataset.LOLv2}, and LOLv2-Synthetic~\cite{Dataset.LOLv2} as the main paired benchmarks to demonstrate the color-faithful restoration capability of our proposed \FRAMEWORK. To further present the effectiveness of our method under more challenging paired low-light conditions, we include more visual comparisons on SDSD-indoor~\cite{Dataset.SDSD}, SDSD-outdoor~\cite{Dataset.SDSD}, and SID~\cite{Dataset.LL.SID}. To evaluate perceptual quality on real low-light scenes without paired normal-light targets, we include DICM~\cite{Dataset.DICM}, LIME~\cite{Dataset.LL.LIME}, MEF~\cite{Dataset.MEF}, NPE~\cite{Dataset.NPE}, and VV~\cite{Dataset.VV}, and adopt the no-reference metrics BRISQUE~\cite{BRISQUE} and NIQE~\cite{NIQE}.

\subsection{More Benchmarking Results}
\label{sec:supp:benchmark}

We provide additional comparisons on both unpaired and paired benchmarks to further present the performance of our method across diverse low-light scenes and imaging conditions, including no-reference quantitative results and visual comparisons on unpaired benchmarks, as well as more qualitative comparisons on paired benchmarks.

\para{More Results on Unpaired Datasets}
As shown in Table~\ref{tab:unpaired_results}, \FRAMEWORK\ improves perceptual quality across different backbone designs and datasets. Over the 15 backbone-dataset combinations built on Retinexformer, DarkIR, and HVI-CIDNet, \FRAMEWORK\ reduces BRISQUE in 12 cases and NIQE in 14 cases. In particular, when integrated into DarkIR, \FRAMEWORK\ reduces BRISQUE by 1.45 / 3.32 / 0.53 / 1.03 / 0.75 and NIQE by 0.124 / 0.301 / 0.020 / 0.105 / 0.130 on DICM, LIME, MEF, NPE, and VV, respectively, and achieves the lowest BRISQUE scores on DICM, LIME, MEF, and NPE. \FRAMEWORK\ further reduces BRISQUE by 7.54 / 3.48 / 8.67 / 3.11 / 10.66 on the five datasets when integrated into HVI-CIDNet, and achieves the best NIQE of 3.249 on MEF together with competitive NIQE results on LIME and VV. 
These results indicate that the improved chromatic modeling in \FRAMEWORK\ consistently benefits no-reference perceptual quality across different backbone designs and datasets, beyond the gains observed in paired full-reference evaluation.

\begin{figure*}[t]
  \centering
  \includegraphics[width=\linewidth]{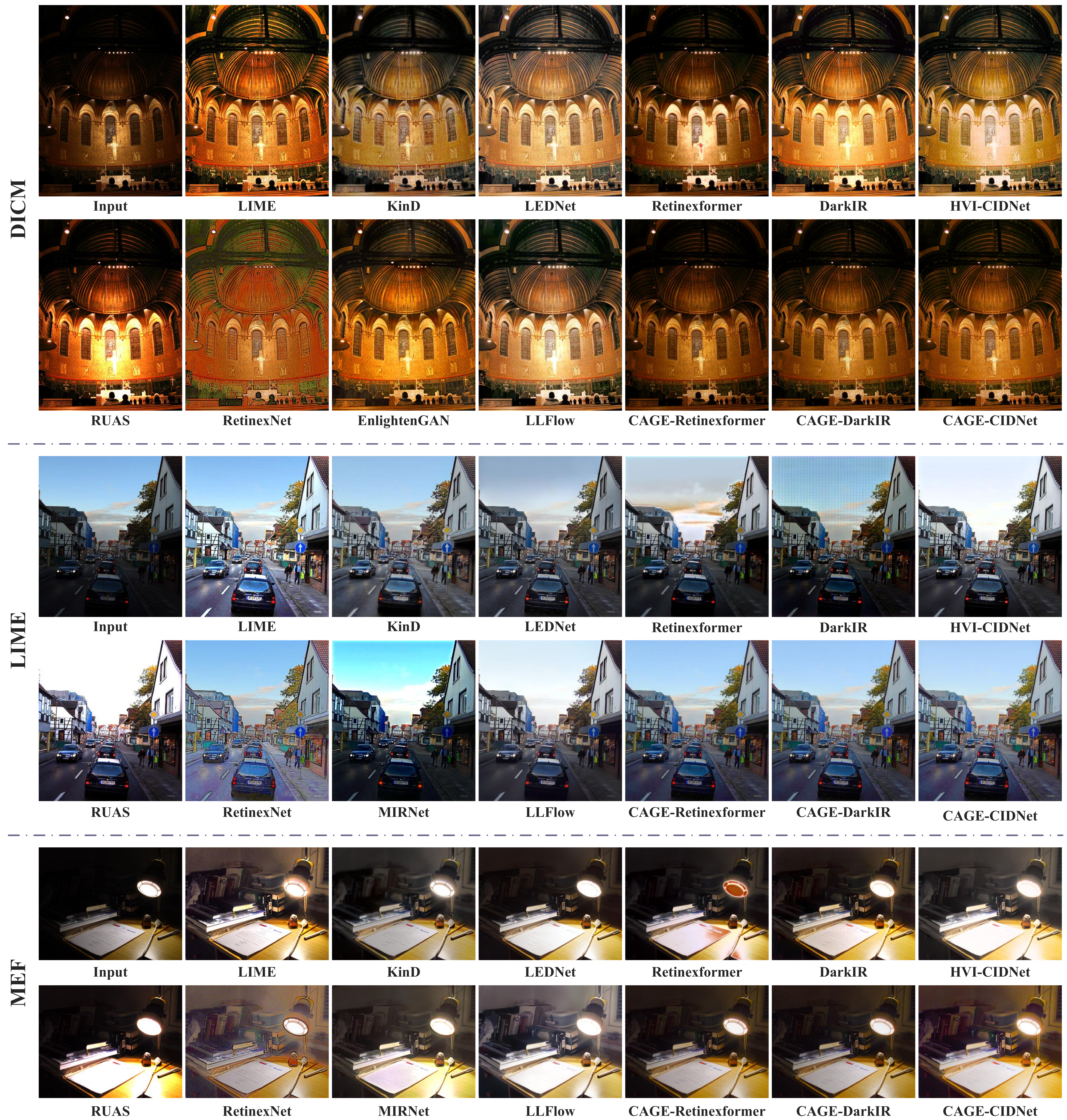}
  \caption{Visual examples on the DICM, LIME, and MEF datasets among LIME~\cite{Dataset.LL.LIME}, KinD~\cite{LL.KinD}, RetinexNet~\cite{LL.RetinexNet.Dataset.LOLv1}, EnlightenGAN~\cite{LL.EnlightenGAN}, MIRNet~\cite{IR.MIRNet}, LEDNet~\cite{IR.LEDNet}, LLFlow~\cite{LL.LLFlow}, and three baseline methods, namely Retinexformer~\cite{LL.Retinexformer}, DarkIR~\cite{LL.DarkIR}, and HVI-CIDNet~\cite{LL.HVI-CIDNet}, with and without our proposed \FRAMEWORK. Our method exhibits stronger exposure balancing and color correction, consequently resulting in more natural brightness, fewer color casts, and clearer local details.}
  \Description{}
  \label{fig:visual-exmaples:unpaired}
\end{figure*}

\begin{figure*}[t]
  \centering
  \includegraphics[width=\linewidth]{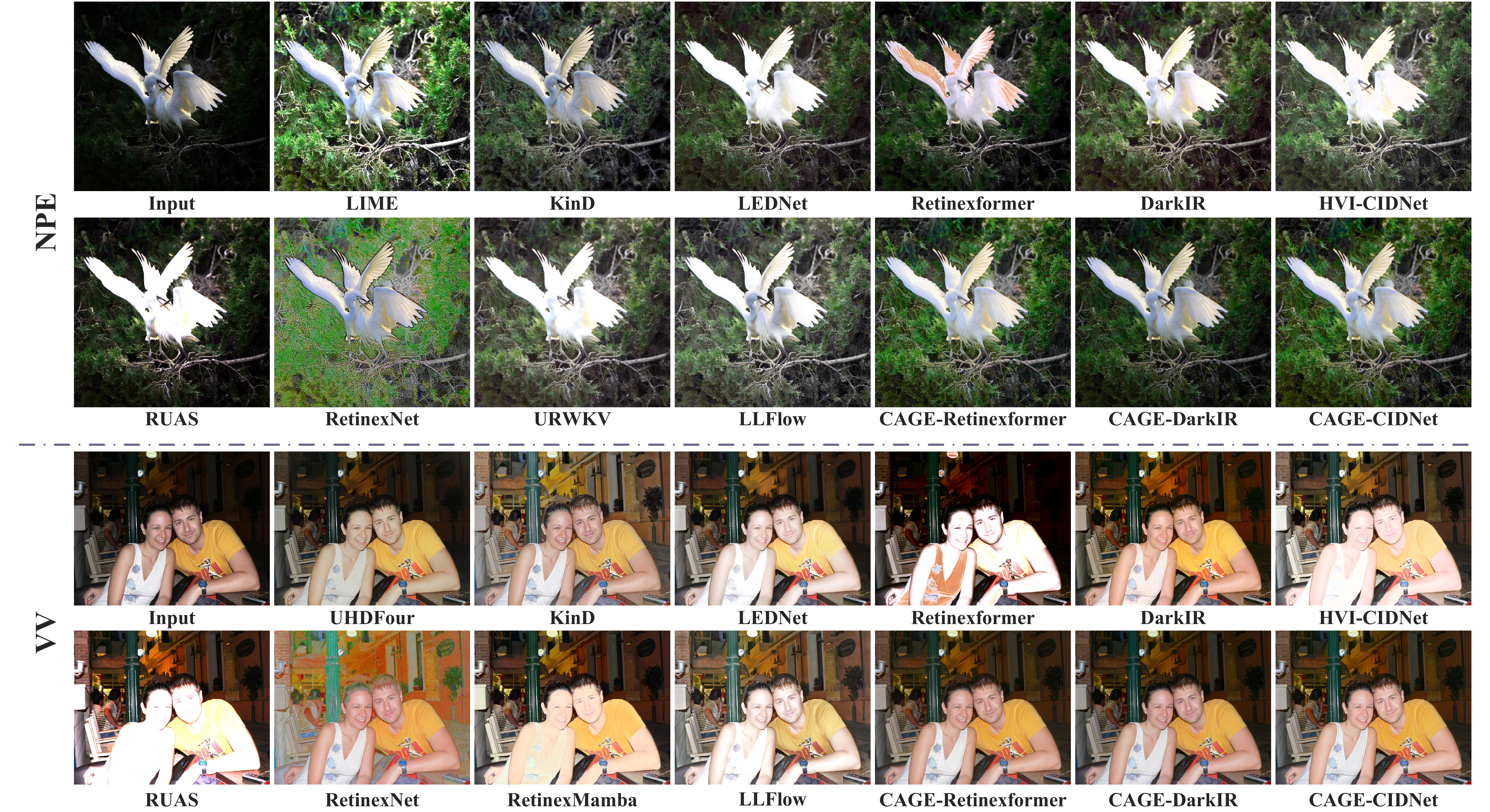}
  \vspace{-6mm}
  \caption{Visual examples on the NPE and VV datasets among UHDFour~\cite{LL.UHDFour}, KinD~\cite{LL.KinD}, RetinexNet~\cite{LL.RetinexNet.Dataset.LOLv1}, RetinexMamba~\cite{LL.RetinexMamba}, LEDNet~\cite{IR.LEDNet}, LLFlow~\cite{LL.LLFlow}, URWKV~\cite{IR.URWKV}, and three baseline methods, namely Retinexformer~\cite{LL.Retinexformer}, DarkIR~\cite{LL.DarkIR}, and HVI-CIDNet~\cite{LL.HVI-CIDNet}, with and without our proposed \FRAMEWORK. Our method exhibits stronger highlight restraint and color recovery, resulting in more balanced exposure, fewer blown-out regions, and more faithful scene colors.}
  \Description{}
  \label{fig:visual-exmaples:unpaired-vv}
\end{figure*}

More visual comparisons on DICM, LIME, and MEF are shown in Fig.~\ref{fig:visual-exmaples:unpaired}, while additional results on NPE and VV are presented in Fig.~\ref{fig:visual-exmaples:unpaired-vv}. These results indicate more natural exposure adjustment, more stable color rendition, and fewer local artifacts under diverse scene illumination.

\begin{figure*}[t]
  \centering
  \includegraphics[width=\linewidth]{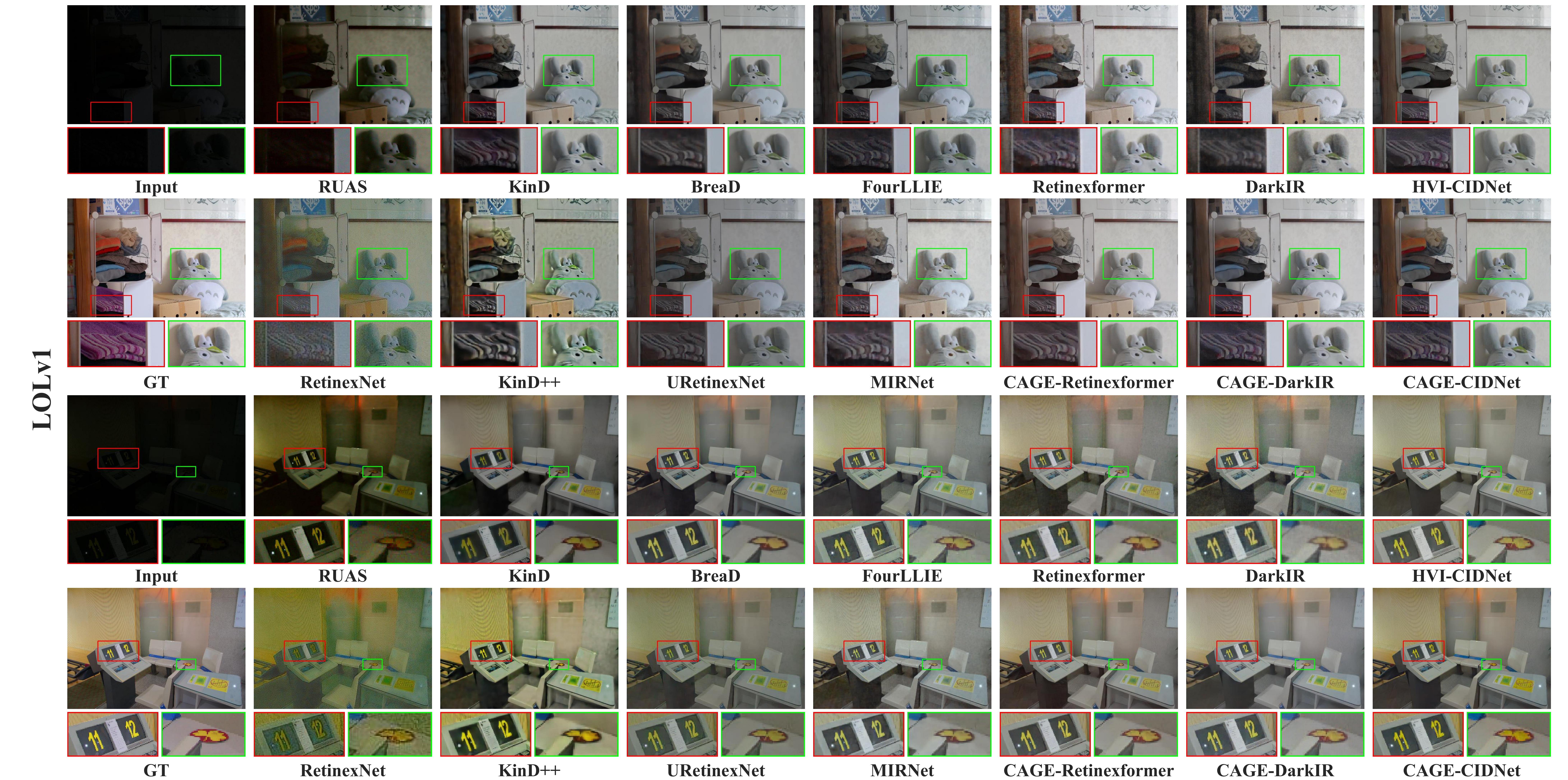}
  \vspace{-6mm}
  \caption{Visual examples on the LOLv1 dataset among RUAS~\cite{LL.RUAS}, KinD~\cite{LL.KinD}, BreaD~\cite{LL.BreaD}, FourLLIE~\cite{LL.FourLLIE}, RetinexNet~\cite{LL.RetinexNet.Dataset.LOLv1}, KinD++~\cite{LL.KinD++}, URetinexNet~\cite{LL.URetinexNet}, and MIRNet~\cite{IR.MIRNet}, as well as  Retinexformer~\cite{LL.Retinexformer}, DarkIR~\cite{LL.DarkIR}, and HVI-CIDNet~\cite{LL.HVI-CIDNet}, with and without our proposed \FRAMEWORK. Our method exhibits better structure-preserving brightness recovery and color correction.}
  \Description{}
  \label{fig:visual-exmaples:LOL}
\end{figure*}

\begin{figure*}[t]
  \centering
  \includegraphics[width=\linewidth]{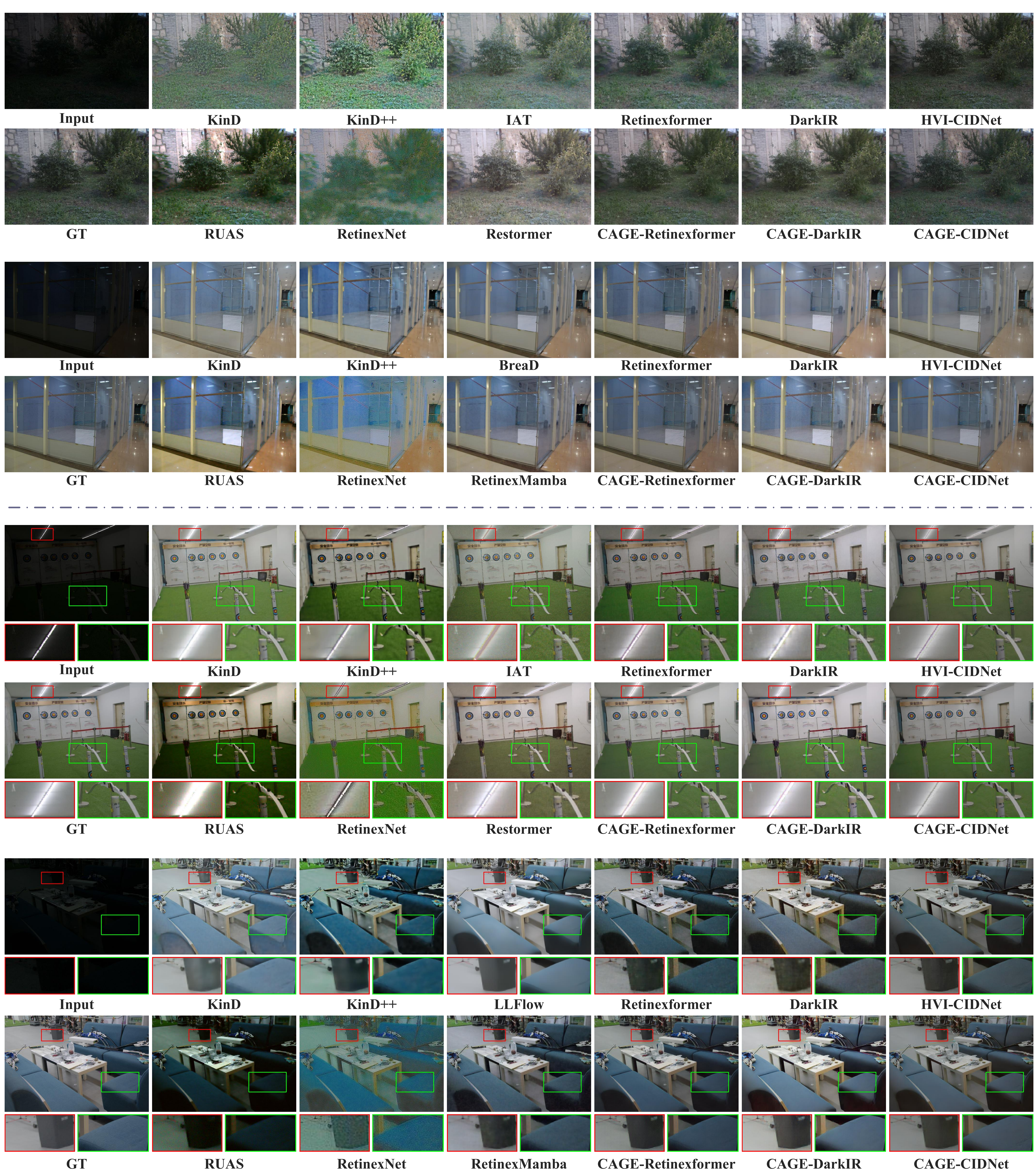}
  \vspace{-6mm}
  \caption{Visual examples on the LOLv2-real dataset among KinD~\cite{LL.KinD}, KinD++~\cite{LL.KinD++}, RetinexNet~\cite{LL.RetinexNet.Dataset.LOLv1}, RUAS, IAT~\cite{LL.IAT}, Restormer~\cite{IR.Restormer}, LLFlow~\cite{LL.LLFlow}, BreaD~\cite{LL.BreaD}, and RetinexMamba~\cite{LL.RetinexMamba}, as well as three baseline methods, namely Retinexformer~\cite{LL.Retinexformer}, DarkIR~\cite{LL.DarkIR}, and HVI-CIDNet~\cite{LL.HVI-CIDNet}, with and without our proposed \FRAMEWORK. Our method yields more balanced brightness restoration and more stable color recovery.}
  \Description{}
  \label{fig:visual-exmaples:LOLv2-real}
\end{figure*}

\begin{figure*}[t]
  \centering
  \includegraphics[width=\linewidth]{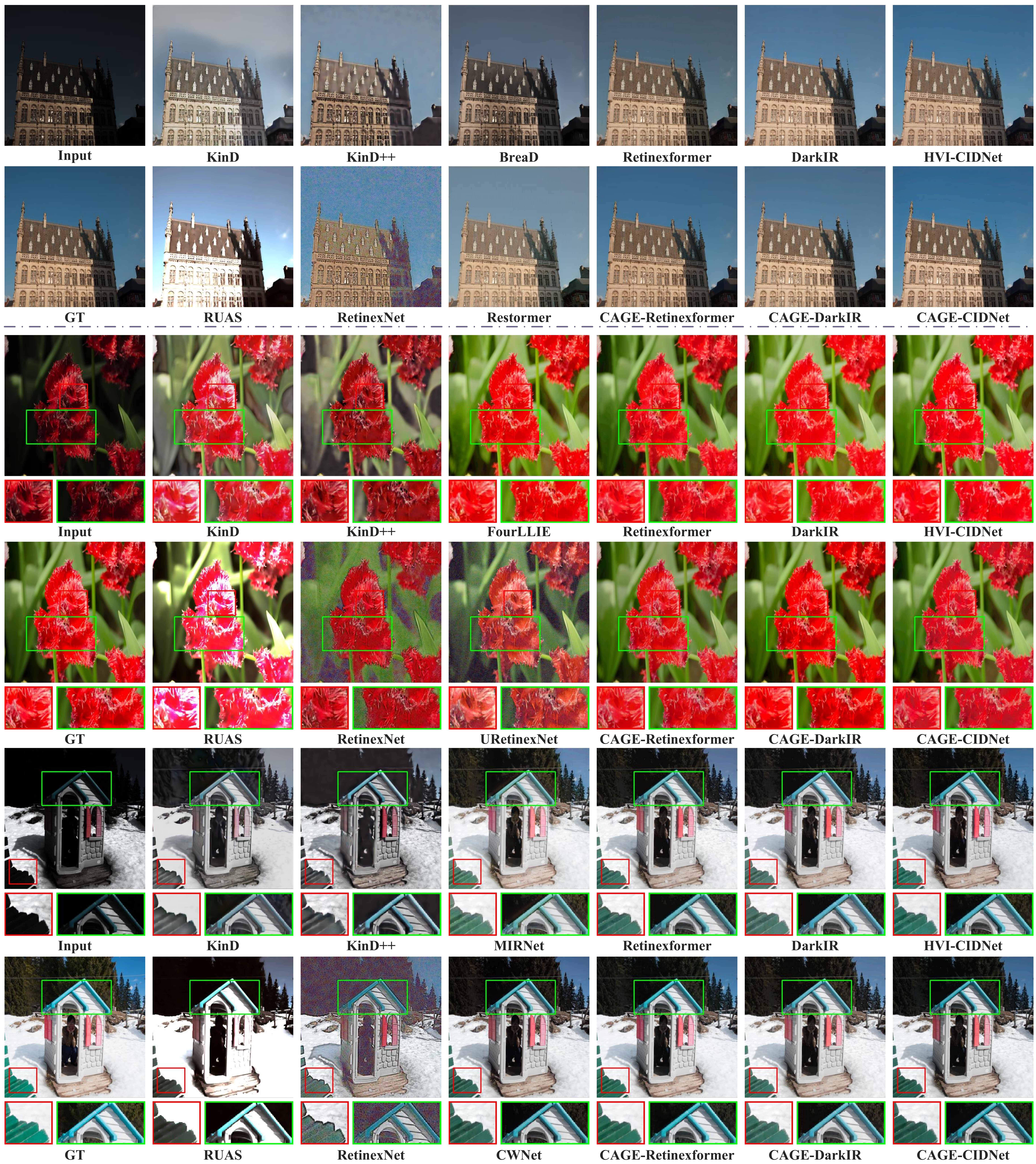}
  \vspace{-6mm}
  \caption{Visual examples for low-light image enhancement on the LOLv2-synthetic dataset among RUAS, KinD~\cite{LL.KinD}, KinD++~\cite{LL.KinD++}, RetinexNet~\cite{LL.RetinexNet.Dataset.LOLv1}, BreaD~\cite{LL.BreaD}, Restormer~\cite{IR.Restormer}, FourLLIE~\cite{LL.FourLLIE}, URetinexNet~\cite{LL.URetinexNet}, MIRNet~\cite{IR.MIRNet}, and CWNet~\cite{LL.CWNet}, as well as three baseline methods, namely Retinexformer~\cite{LL.Retinexformer}, DarkIR~\cite{LL.DarkIR}, and HVI-CIDNet~\cite{LL.HVI-CIDNet}, with and without our proposed \FRAMEWORK. Our method yields more balanced exposure and more faithful color restoration, consequently resulting in outputs with clearer local structures and fewer artifacts.}
  \Description{}
  \label{fig:visual-exmaples:LOLv2-synthetic}
\end{figure*}

\begin{figure*}[t]
  \centering
  \includegraphics[width=\linewidth]{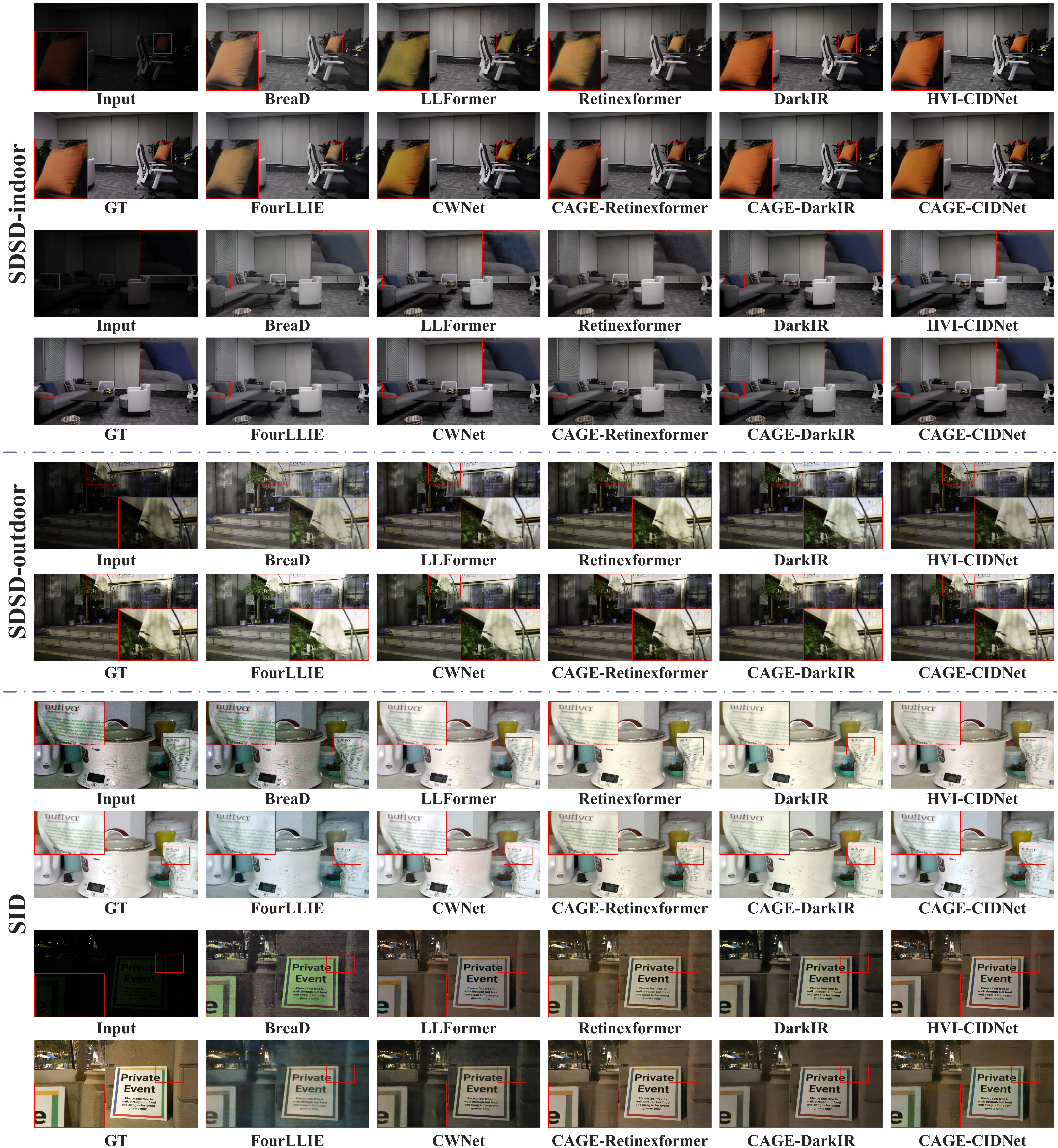}
  \vspace{-6mm}
  \caption{Visual examples for low-light image enhancement on the SDSD-indoor, SDSD-outdoor, and SID datasets among BreaD~\cite{LL.BreaD}, LLFormer~\cite{LL.LLFormer}, FourLLIE~\cite{LL.FourLLIE}, and CWNet~\cite{LL.CWNet}, as well as three baseline methods, namely Retinexformer~\cite{LL.Retinexformer}, DarkIR~\cite{LL.DarkIR}, and HVI-CIDNet~\cite{LL.HVI-CIDNet}, with and without our proposed \FRAMEWORK. In SDSD-indoor, \FRAMEWORK\ recovers severely dark regions while maintaining a smooth intensity transition, avoiding the over-flattened appearance. In SDSD-outdoor, it preserves local contrast under strong illumination variation, preventing detail loss around bright structures. In SID, \FRAMEWORK\ enhances fine textual content and small objects with clearer edges, reducing color contamination and structural ambiguity in challenging low-light conditions.}
  \Description{}
  \label{fig:visual-exmaples:SDSD+SID}
\end{figure*}

\para{More Qualitative Comparisons on Paired Datasets}
More qualitative comparisons on LOLv1~\cite{LL.RetinexNet.Dataset.LOLv1}, LOLv2-Real~\cite{Dataset.LOLv2}, LOLv2-Synthetic~\cite{Dataset.LOLv2}, SDSD-indoor~\cite{Dataset.SDSD}, SDSD-outdoor~\cite{Dataset.SDSD}, and SID~\cite{Dataset.LL.SID} are shown in Figs.~\ref{fig:visual-exmaples:LOL}--\ref{fig:visual-exmaples:SDSD+SID}. On these paired datasets, our method produces reconstructions with more faithful brightness restoration, cleaner structural details, and better color consistency with the reference images.

\subsection{More Ablation Studies}
\label{sec:supp:abl}
We further include more ablation experiments on the real-world LOL-v2-real dataset to assess the contribution of each component in \FRAMEWORK\ and provide a more detailed analysis of their effects on color-faithful low-light image enhancement.

\begin{figure}[t]
  \centering
  \includegraphics[width=\linewidth]{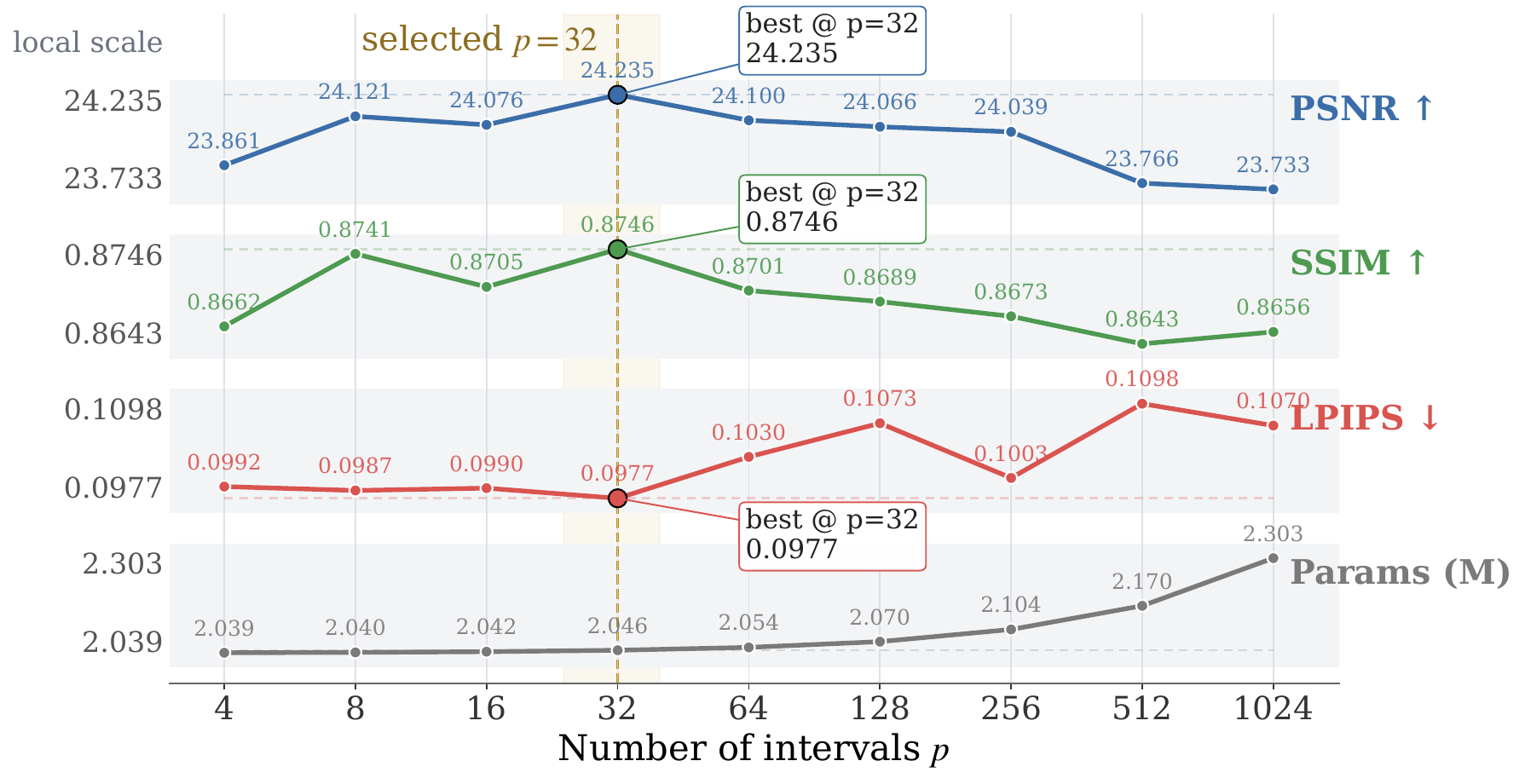}
  \vspace{-8mm}
  \caption{Ablation of the interval number $p$.}
  \Description{}
  \label{fig:abl:intervals_numbers}
\end{figure}

\para{Number of Intervals $p$}
We vary the number of intervals $p$ to analyze its effect on adaptive lightness sensitivity modeling. As shown in Fig.~\ref{fig:abl:intervals_numbers}, the overall performance improves as $p$ increases from 4 to 32, and then drops when $p$ becomes larger. The best setting is $p=32$, which achieves 24.2355 dB PSNR, 0.8746 SSIM, and 0.0977 LPIPS. A small $p$ gives an overly coarse partition on the lightness axis, while an excessively large $p$ produces overly fine intervals and reduces the robustness of interpolation. Meanwhile, the parameter count increases only slightly from 2.039M to 2.303M as $p$ increases from 4 to 1024. Therefore, we use $p=32$ as the default setting.

\para{Global Interaction in Vertex Values}
We further compare four strategies for learning the vertex values on the lightness axis, including the sinusoidal curve used in HVI-CIDNet~\cite{LL.HVI-CIDNet}, a plain 1D look-up-table (1DLUT)~\cite{CT.1DLUT}, a 1D LUT with monotonicity, and the proposed 1D LUT with global interaction. As shown in Table~\ref{tab:abl:gamut}, replacing the sinusoidal curve with a plain 1D LUT improves PSNR from 23.9560 to 24.1666, SSIM from 0.8701 to 0.8727, and reduces LPIPS from 0.1080 to 0.1015. This result shows that the vertex values in LAB require a more flexible lightness-wise parameterization than a single smooth curve. Adding monotonicity to the 1D LUT further reduces PSNR to 23.7778 and gives an inferior LPIPS of 0.1024. This behavior is consistent with the non-monotonic feasible chroma range of LAB along the lightness axis, where a monotonic parameterization cannot follow the lightness-dependent chromatic variation well. The proposed 1D LUT with global interaction achieves the best results, reaching a PSNR of 24.2355 dB, 0.8746 SSIM, and 0.0977 LPIPS. Compared with the plain 1D LUT, global interaction further improves PSNR by 0.0689, SSIM by 0.0019, and reduces LPIPS by 0.0038. The gain comes from allowing each vertex value to interact with all lightness vertices rather than only two adjacent samples, which yields more consistent chromatic reorganization across lightness levels.

\input{tables/supp-ablation.tex}

\subsection{Efficiency Analysis}
\label{sec:supp:efficiency_overhead}

To further quantify the efficiency overhead introduced by \FRAMEWORK, we compare Retinexformer, DarkIR, and HVI-CIDNet with and without \FRAMEWORK\ in terms of parameters, FLOPs, latency, and memory usage. Following the experimental environment described in Sec.~\ref{sec:supp:exp_setup}, latency and memory usage are measured on a single NVIDIA A40 GPU.

As shown in Table~\ref{tab:supp:efficiency_overhead}, \FRAMEWORK\ adds \textbf{0.070M} parameters and \textbf{0.005G} FLOPs to all three backbones (the FLOPs are shown to three decimal places to make this small increment visible). Consistent with the analysis in the main body, the extra cost is very small, indicating that the performance gains mainly comes from improved color modeling rather than increased model capacity. Across the different resolutions of the representative method (\ie, HVI-CIDNet), the additional latency remains below \textbf{9 ms}, while the memory overhead stays close to \textbf{7\%}, showing that the resolution-dependent efficiency overhead is mainly determined by the LLIE backbone rather than \FRAMEWORK.

\section{Human Subjective Evaluation}
\label{sec:supp:human_subjective_evaluation}

\begin{figure}[t]
  \centering
  \includegraphics[width=\linewidth]{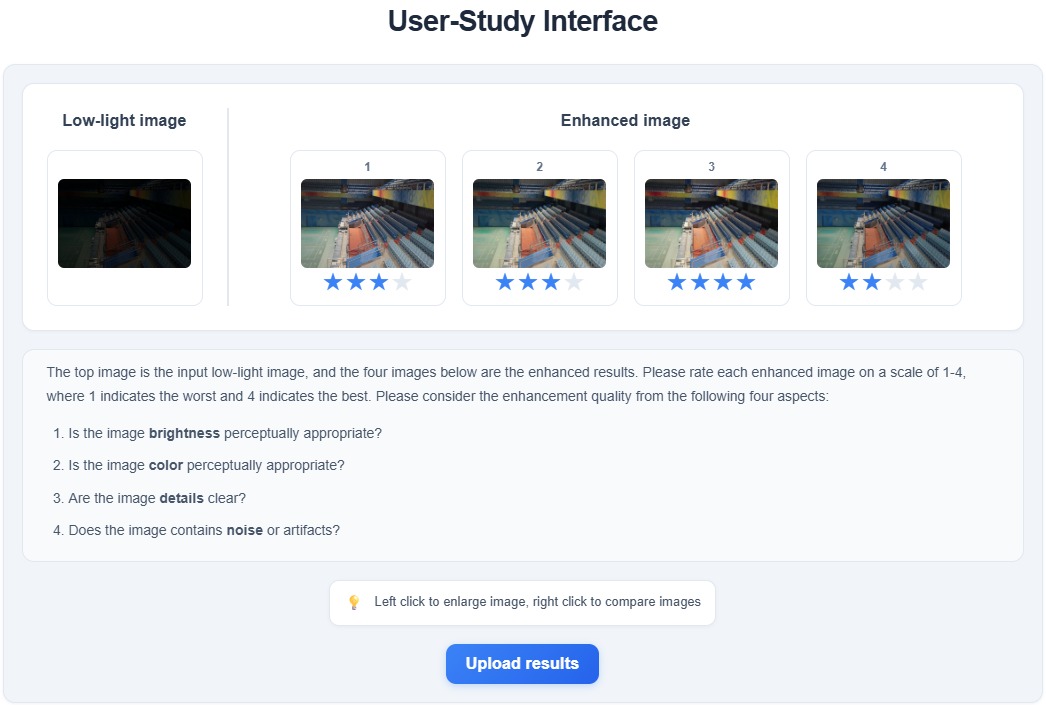}
  \vspace{-6mm}
  \caption{Interface used for human subjective evaluation.}
  \label{fig:supp:user_interface}
\end{figure}

To further assess the perceptual improvement brought by \FRAMEWORK, we include a human subjective evaluation with 12 participants and more than 1,500 individual ratings. For each low-light input, one result is randomly selected from \FRAMEWORK-integrated methods (\ie, \FRAMEWORK-CIDNet, \FRAMEWORK-Retinexformer, and \FRAMEWORK-DarkIR), and three results are randomly selected from the comparison methods. The four results are displayed in random order without method names, reducing selection and position bias. As shown in Fig.~\ref{fig:supp:user_interface}, participants score each result from 0 to 4 based on brightness, color, detail clarity, and noise or artifacts. The interface supports image enlargement and freely selected comparisons, allowing participants to inspect the overall enhancement and local details.

\begin{figure}[t]
  \centering
  \includegraphics[width=\linewidth]{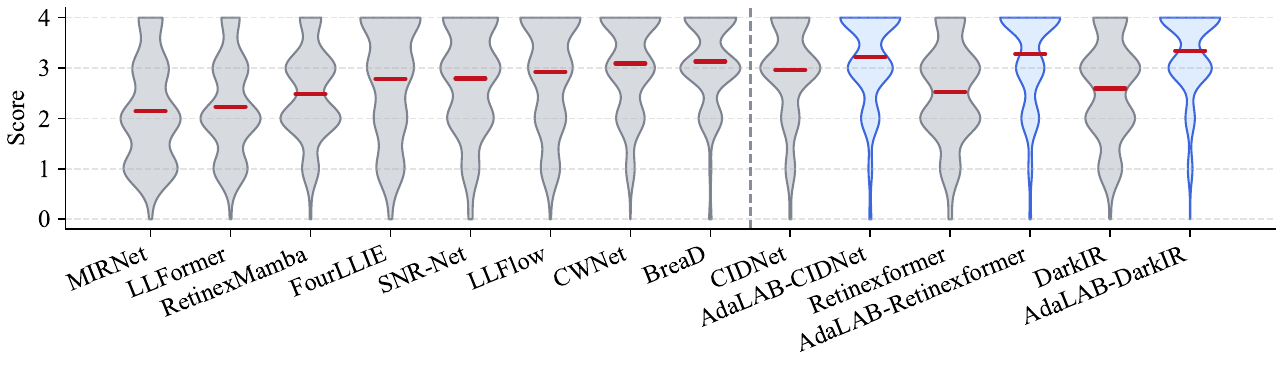}
  \vspace{-6mm}
  \caption{Distribution of subjective scores for representative LLIE methods and three backbones w/ and w/o \FRAMEWORK. Red horizontal lines indicate the mean scores. Scores range from 0 to 4, with higher scores indicating better perceptual quality.}
  \label{fig:supp:human_subjective_evaluation}
\end{figure}

As shown in Fig.~\ref{fig:supp:human_subjective_evaluation}, integrating \FRAMEWORK\ increases the mean subjective scores of three backbones (\eg, HVI-CIDNet, Retinexformer, and DarkIR), showing that the perceptual gain of \FRAMEWORK\ is not restricted to a specific backbone design. Among the comparison methods, BreaD receives the highest mean score, while its outputs display noticeably higher saturation. Nevertheless, the \FRAMEWORK\ variants still surpass BreaD, indicating that the perceptual advantage of \FRAMEWORK\ extends beyond the visual appeal associated with pronounced saturation.

\begin{table}[t]
    \centering
    \caption{Efficiency overhead of LLIE backbones + \FRAMEWORK.}
    \vspace{-3mm}
    \label{tab:supp:efficiency_overhead}
    \setlength{\tabcolsep}{0.8pt}
    \renewcommand{\arraystretch}{0.8}
    \resizebox{\linewidth}{!}{
        \begin{tabular}{cc!{\vrule}c!{\vrule}llll}
        \toprule
        Backbone & Resolution & Variant & Params (M) & FLOPs (G) & Latency (ms) & Memory (MB) \\
        \midrule
        \multirow{2}{*}{Retinexformer}
        & \multirow{2}{*}{$256\times256$}
        & Base & 1.606 & \, 17.019 & \, 19.42 $\pm$1.43 & \, 1358 \\
        & & \FRAMEWORK & 1.676 \gt{0.070} & \, 17.024 \gt{0.005} & \, 24.99 $\pm$0.17 \gt{5.57} & \, 1462 \gt{7.7\%} \\
        \midrule
        \multirow{2}{*}{DarkIR}
        & \multirow{2}{*}{$256\times256$}
        & Base & 3.322 & \, 7.110 & \, 22.94 $\pm$0.25 & \, 960 \\
        & & \FRAMEWORK & 3.392 \gt{0.070} & \, 7.115 \gt{0.005} & \, 31.80 $\pm$0.24 \gt{8.86} & \, 1020 \gt{6.3\%} \\
        \midrule
        \multirow{6}{*}{HVI-CIDNet}
        & \multirow{2}{*}{$256\times256$}
        & Base & 1.976 & \, 8.133 & \, 27.66 $\pm$0.21 & \, 906 \\
        & & \FRAMEWORK & 2.046 \gt{0.070} & \, 8.138 \gt{0.005} & \, 34.34 $\pm$0.19 \gt{6.68} & \, 978 \gt{7.9\%} \\
        \cmidrule{2-7}
        & \multirow{2}{*}{$512\times512$}
        & Base & 1.976 & \, 32.532 & \, 79.80 $\pm$0.39 & \, 3610 \\
        & & \FRAMEWORK & 2.046 \gt{0.070} & \, 32.537 \gt{0.005} & \, 86.63 $\pm$0.28 \gt{6.83} & \, 3876 \gt{7.4\%} \\
        \cmidrule{2-7}
        & \multirow{2}{*}{$1024\times1024$}
        & Base & 1.976 & 130.127 & 308.11 $\pm$0.65 & 13870 \\
        & & \FRAMEWORK & 2.046 \gt{0.070} & 130.132 \gt{0.005} & 316.83 $\pm$0.22 \gt{8.72} & 14966 \gt{7.9\%} \\
        \bottomrule
        \end{tabular}
    }

    \begin{minipage}{\linewidth}
    \footnotesize
    \raggedright
    Latency denotes mean $\pm$ std over 100 runs. Green values denote the overhead introduced by \FRAMEWORK.
    \end{minipage}
\end{table}

\section{Limitations}

Although \FRAMEWORK\ improves color-faithful low-light image enhancement by explicitly modeling chromatic disturbance in the \LAB\ space, the shared hue-shift direction defines its main application boundary. Hue-directional chromatic debiasing predicts one image-level shift direction and varies its magnitude along the lightness axis. This formulation matches low-light color bias whose direction is coherent across the image and whose magnitude changes with lightness. Under mixed illumination, however, spatially separated regions with similar lightness may exhibit different color-shift directions. Since the current formulation assigns the same direction to these regions, residual local color bias may persist after enhancement. A future extension can incorporate local chromatic content into direction prediction, allowing regions with similar lightness to receive different hue-shift directions.

\end{document}

%% file: tables/main-lol_results.tex
\begin{table*}[t]
  \centering
  \setlength{\tabcolsep}{0.55mm}       
  \renewcommand{\arraystretch}{0.92}    
  \caption{Quantitative comparisons on LOLv1, LOLv2-real, and LOLv2-synthetic datasets.}
  \vspace{-2mm}
  \resizebox{\linewidth}{!}{
    \begin{tabular}{l!{\vrule}cc!{\vrule}ccc!{\vrule}ccc!{\vrule}ccc}
    \toprule
    \multirow{2}[2]{*}{\textbf{Methods}}
      & \multicolumn{2}{c!{\vrule}}{\textbf{Complexity}}
      & \multicolumn{3}{c!{\vrule}}{\textbf{LOLv1}}
      & \multicolumn{3}{c!{\vrule}}{\textbf{LOLv2-real}}
      & \multicolumn{3}{c}{\textbf{LOLv2-synthetic}} \\
      & Params (M) & FLOPs (G)
      & PSNR  \up & SSIM  \up & LPIPS  \down
      & PSNR  \up & SSIM  \up & LPIPS  \down
      & PSNR  \up & SSIM  \up & LPIPS  \down\\
    \midrule

    RetinexNet~\cite{LL.RetinexNet.Dataset.LOLv1} {\scriptsize BMVC'18}
      & \, \, \, 0.84 & \, 584.47
      & 17.56 & 0.645 & 0.378
      & 17.68 & 0.642 & 0.440
      & 15.61 & 0.449 & 0.746 \\

    KinD~\cite{LL.KinD} {\scriptsize ACM MM'19}
      & \, \, \, 8.02 & \, \, 64.73
      & 17.64 & 0.777 & 0.175
      & 20.59 & 0.818 & 0.143
      & 16.26 & 0.591 & 0.435 \\

    DRBN~\cite{LL.DRBN} {\scriptsize CVPR'20}
      & \, \, \, 5.27 & \, \, 48.61
      & 20.13 & 0.830 & 0.155
      & 20.29 & 0.831 & 0.147
      & 23.22 & 0.927 & - \\

    MIRNet~\cite{IR.MIRNet} {\scriptsize ECCV'20}
    & \, \, 31.76 & \, 785.00
    & 26.56 & 0.853 & 0.128
    & 22.68 & 0.828 & 0.212
    & 25.05 & 0.923 & 0.072 \\

    ZeroDCE~\cite{LL.Zero-DCE} {\scriptsize CVPR'20}
      & \, \, \, 0.08 & \, \, 19.01
      & 21.88 & 0.640 & 0.335
      & 16.06 & 0.580 & 0.313
      & 17.71 & 0.815 & 0.169 \\

    EnlightenGAN~\cite{LL.EnlightenGAN} {\scriptsize TIP'21}
      & \, 114.35 & \, \, 61.01
      & 20.00 & 0.691 & 0.317
      & 18.23 & 0.617 & 0.309
      & 16.57 & 0.734 & 0.220 \\

    LLFlow~\cite{LL.LLFlow} {\scriptsize AAAI'22}
      & \, \, 17.42 & \, 358.4
      & 25.53 & \second{0.870} & 0.111
      & 17.43 & 0.831 & 0.176
      & 23.43 & 0.933 & 0.050 \\

    SNR-Net~\cite{LL.SNRNet} {\scriptsize CVPR'22}
      & \, \, \, 4.01 & \, \, 26.35
      & 26.81 & 0.855 & 0.150
      & 21.48 & 0.849 & 0.157
      & 24.14 & 0.927 & 0.056 \\

    BreaD~\cite{LL.BreaD} {\scriptsize IJCV'23}
      & \, \, \, 2.02 & \, \, 19.85
      & 25.32 & 0.843 & 0.155
      & 23.69 & 0.869 & 0.156
      & 15.97 & 0.746 & 0.261 \\

    FourLLIE~\cite{LL.FourLLIE} {\scriptsize ACM MM'23}
      & \, \, \, 0.12 & -
      & 25.05 & 0.839 & 0.121
      & 22.35 & 0.847 & 0.114
      & 24.65 & 0.919 & 0.066 \\

    LLFormer~\cite{LL.LLFormer} {\scriptsize AAAI'23}
      & \, \, 24.52 & \, \, 22.52
      & 26.12 & 0.827 & 0.166
      & 21.65 & 0.816 & 0.205
      & 25.26 & 0.923 & 0.061 \\

    QuadPrior~\cite{LL.QuadPrior} {\scriptsize CVPR'24}
      & 1252.75 & 1103.20
      & 22.85 & 0.800 & 0.201
      & 20.59 & 0.811 & 0.202
      & 16.11 & 0.758 & 0.114 \\

    CWNet~\cite{LL.CWNet} {\scriptsize ICCV'25}
      & \, \, \, 1.23 & \, \, 11.30
      & 26.38 & 0.864 & 0.120
      & 21.65 & 0.860 & 0.136
      & 25.50 & \second{0.937} & \second{0.044} \\

    \midrule

    Retinexformer~\cite{LL.Retinexformer} {\scriptsize ICCV'23}
      & \, \, \, 1.61 & \, \, 17.02
      & 26.08 & 0.827 & 0.155 
      & 21.89 & 0.837 & 0.159
      & 24.59 & 0.919 & 0.069 \\

    \oursrow
      & \, \, \, 1.68 & \, \, 17.03
      & \second{27.09} \gt{1.01} & \first{0.871} \gt{0.044} & \second{0.090} \gt{0.065}
      & \second{24.10} \gt{2.21} & 0.847 \gt{0.010} & \second{0.104} \gt{0.055}
      & \first{26.33} \gt{1.74} & \first{0.939} \gt{0.020} & \first{0.039} \gt{0.030} \\
    \midrule

    DarkIR~\cite{LL.DarkIR} {\scriptsize CVPR'25}
      & \, \, \, 3.32 & \, \, \, 7.11
      & 25.32 & 0.806 & 0.106
      & 21.32 & 0.830 & 0.139
      & 24.06 & 0.921 & 0.060 \\

    \oursrow
      & \, \, \, 3.39 & \, \, \, 7.12
      & 26.03 \gt{0.71} & 0.849 \gt{0.043} & 0.106 \gt{0.000}
      & 22.51 \gt{1.19} & 0.860 \gt{0.030} & 0.121 \gt{0.018}
      & 24.54 \gt{0.48} & 0.923 \gt{0.002} & 0.058 \gt{0.002} \\
    \midrule

    HVI-CIDNet~\cite{LL.HVI-CIDNet} {\scriptsize CVPR'25}
      & \, \, \, 1.98 & \, \, \, 8.13
      & 26.03 & 0.832 & 0.101
      & 23.94 & \second{0.870} & 0.113
      & 25.32 & 0.929 & 0.054 \\

    \oursrow
      & \, \, \, 2.05 & \, \, \, 8.14
      & \first{27.25} \gt{1.22} & 0.857 \gt{0.025} & \first{0.083} \gt{0.018}
      & \first{24.24} \gt{0.30} & \first{0.875} \gt{0.005} & \first{0.098} \gt{0.015}
      & \second{25.52} \gt{0.20} & 0.930 \gt{0.001} & 0.046 \gt{0.008} \\
    \bottomrule
    \end{tabular}
  }

  \begin{minipage}{\linewidth}
    \footnotesize
    \raggedright
    The best and second-best results are highlighted in \first{bold} and \second{underline}, respectively.
    \gtsign{+}(\lssign{-}) denotes the \gtsign{improvement} (\lssign{reduction}) of performance, corresponding to $\uparrow$ ($\downarrow$).
  \end{minipage}
  \label{tab:main_results_lol}
\end{table*}

%% file: tables/main-sdsd_results.tex
\begin{table}[t]
  \centering
  \setlength{\tabcolsep}{0.3mm}
  \renewcommand{\arraystretch}{1.0}    
  \caption{Quantitative comparisons on challenging datasets.}
  \resizebox{\linewidth}{!}{
    \begin{tabular}{l!{\vrule}ccc!{\vrule}ccc!{\vrule}ccc}
    \toprule
    \multirow{2}[2]{*}{\textbf{Methods}}
      & \multicolumn{3}{c!{\vrule}}{\textbf{SDSD-indoor}}
      & \multicolumn{3}{c!{\vrule}}{\textbf{SDSD-outdoor}}
      & \multicolumn{3}{c}{\textbf{SID}} \\
      & PSNR & SSIM & LPIPS
      & PSNR & SSIM & LPIPS
      & PSNR & SSIM & LPIPS \\
    \midrule

    3DLUT~\cite{CT.3DLUT}
      & 24.78 & 0.718 & -
      & 23.29 & 0.703 & -
      & 16.97 & 0.591 & - \\

    SNR-Net~\cite{LL.SNRNet}
      & 26.13 & 0.815 & 0.198
      & 19.22 & 0.657 & 0.225
      & 21.35 & 0.550 & 0.446 \\

    LEDNet~\cite{IR.LEDNet}
      & 27.29 & 0.876 & 0.121
      & 26.66 & 0.850 & 0.145
      & 21.47 & 0.638 & 0.349 \\

    FourLLIE~\cite{LL.FourLLIE}
      & 24.74 & 0.826 & 0.146
      & 24.67 & 0.787 & 0.158
      & 18.42 & 0.513 & 0.504 \\

    RetinexMamba~\cite{LL.RetinexMamba}
      & 28.44 & 0.894 & 0.119
      & 28.52 & 0.859 & 0.180
      & 22.45 & 0.656 & 0.351 \\

    MIRNet~\cite{IR.MIRNet}
      & 28.64 & 0.888 & 0.128
      & 28.99 & 0.869 & 0.152
      & 21.36 & 0.632 & 0.397 \\

    Restormer~\cite{IR.Restormer}
      & 28.49 & 0.892 & 0.124
      & 27.99 & 0.868 & 0.189
      & 22.01 & 0.645 & 0.361 \\

    MambaIR~\cite{IR.MambaIR}
      & 25.14 & 0.876 & 0.130
      & 27.53 & 0.851 & 0.177
      & 22.02 & \second{0.658} & 0.337 \\

    CWNet~\cite{LL.CWNet}
      & \second{30.28} & \second{0.904} & 0.127
      & 28.54 & 0.857 & 0.154
      & 20.97 & 0.623 & 0.429 \\
    \midrule

    Retinexformer~\cite{LL.Retinexformer}
      & 28.20 & 0.877 & 0.137
      & 29.57 & \second{0.874} & 0.122
      & 22.05 & 0.636 & 0.372 \\

    \oursrow
      & 29.98 & \first{0.906} & \first{0.072}
      & \first{30.12} & \first{0.877} & \second{0.100}
      & \first{22.64} & \first{0.660} & \first{0.281} \\
    \deltarow & \gt{1.78} & \gt{0.029} & \gt{0.065}
      & \gt{0.55} & \gt{0.003} & \gt{0.022}
      & \gt{0.59} & \gt{0.024} & \gt{0.091} \\
    \midrule

    DarkIR~\cite{LL.DarkIR}
      & 29.94 & 0.897 & 0.096
      & 28.81 & 0.848 & 0.104
      & 21.50 & 0.588 & 0.469 \\

    \oursrow
      & \first{30.36} & 0.899 & \second{0.080}
      & 29.36 & 0.858 & 0.101
      & 22.33 & 0.639 & 0.305 \\
    \deltarow & \gt{0.42} & \gt{0.002} & \gt{0.016}
      & \gt{0.55} & \gt{0.010} & \gt{0.003}
      & \gt{0.83} & \gt{0.051} & \gt{0.164} \\
    \midrule

    HVI-CIDNet~\cite{LL.HVI-CIDNet}
      & 28.74 & 0.895 & 0.090
      & 28.70 & 0.866 & 0.107
      & 22.21 & 0.631 & 0.308 \\

    \oursrow
      & 29.91 & 0.900 & 0.083
      & \second{29.80} & \first{0.877} & \first{0.097}
      & \second{22.46} & 0.653 & \second{0.286} \\
    \deltarow & \gt{1.17} & \gt{0.005} & \gt{0.007}
      & \gt{1.10} & \gt{0.011} & \gt{0.010}
      & \gt{0.25} & \gt{0.022} & \gt{0.022} \\
    \bottomrule
    \end{tabular}
  }

  \begin{minipage}{\linewidth}
    \footnotesize
    \raggedright
    The best and second-best results are highlighted in \first{bold} and \second{underline}, respectively.
    \gtsign{+}(\lssign{-}) denotes the \gtsign{improvement} (\lssign{reduction}) of performance, corresponding to $\uparrow$ ($\downarrow$).
  \end{minipage}
  \label{tab:main_results_sdsd_sid}
  \vspace{-4mm}
\end{table}

%% file: tables/main-ablation.tex
\begin{table}[t]
  \vspace{-2mm}
  \centering
  \renewcommand{\arraystretch}{0.8}    
  \caption{Ablation of image-adaptive cylindrical parameters.}
  \vspace{-2mm}
  \label{tab:abl:cylindrical}
  \setlength{\tabcolsep}{6pt}
  \renewcommand{\arraystretch}{1.12}
  \resizebox{\linewidth}{!}
  {
    \begin{tabular}{c!{\vrule}ccc!{\vrule}ccc}
    \toprule
    \rowcolor{gray}
    \# & \underline{H}ue P. & \underline{C}hroma P. & \underline{L}ightness P. & PSNR \up & SSIM \up & LPIPS \down \\
    \midrule
    \circnum{1} &        & \cmark & \cmark & 23.7493 & 0.8639 & 0.0980 \\
    \circnum{2} & \cmark &        & \cmark & 24.0237 & 0.8734 & 0.1046 \\
    \circnum{3} & \cmark & \cmark &        & 24.2201 & 0.8667 & 0.1006 \\
    \circnum{4} & \cmark & \cmark & \cmark & \textbf{24.2355} & \textbf{0.8746} & \textbf{0.0977} \\
    \bottomrule
    \end{tabular}
  }

  \vspace{1mm}
  \begin{minipage}{\linewidth}
    \footnotesize
    \raggedright
    Hue P., Chroma P., and Lightness P. denote Hue-shift Direction and Magnitude, Chroma-scaling Intensity, and Lightness Sensitivity Vertices, respectively.
  \end{minipage}
\end{table}

\begin{table}[t]
  \vspace{-2mm}
  \centering
  \caption{Ablation of color space design.}
  \vspace{-2mm}
  \label{tab:abl:color_space}
  \setlength{\tabcolsep}{9pt}
  \renewcommand{\arraystretch}{1.12}
  \resizebox{\linewidth}{!}{
    \begin{tabular}{c!{\vrule}>{\centering\arraybackslash}p{3.2cm}!{\vrule}ccc}
    \toprule
    \rowcolor{gray}
    \# & Color Space & PSNR \up & SSIM \up & LPIPS \down \\
    \midrule
    \circnum{1} & RGB color space & 22.3870 & 0.8334 & 0.1585 \\
    \circnum{2} & YUV color space & 23.2105 & 0.8630 & 0.1117 \\
    \circnum{3} & HVI color space & 23.9364 & 0.8712 & 0.1128 \\
    \circnum{4} & CIELab color space & 24.0171 & 0.8718 & 0.1049 \\
    \circnum{5} & AdaLAB color space & \textbf{24.2355} & \textbf{0.8746} & \textbf{0.0977} \\
    \bottomrule
    \end{tabular}
  }
\end{table}

\begin{table}[t]
  \vspace{-2mm}
  \centering
  \caption{Ablation of gamut harmonization on AdaLAB.}
  \vspace{-2mm}
  \label{tab:abl:gamut}
  \setlength{\tabcolsep}{5pt}
  \renewcommand{\arraystretch}{1.12}
  \resizebox{\linewidth}{!}
  {
    \begin{tabular}{c!{\vrule}c!{\vrule}ccc}
    \toprule
    \rowcolor{gray}
    \# & Gamut Handling & PSNR \up & SSIM \up & LPIPS \down \\
    \midrule
    \circnum{1} & RGB gamut clipping  & 23.4189 & 0.8640 & 0.1002 \\
    \circnum{2} & LAB gamut clipping  & 24.2272 & 0.8689 & 0.0996 \\
    \circnum{3} & Out-of-gamut lightness compensation  & \textbf{24.2355} & \textbf{0.8746} & \textbf{0.0977} \\
    \bottomrule
    \end{tabular}
  }
\end{table}

%% file: tables/supp-unpaired_results.tex
\begin{table*}[t]
  \centering
  \setlength{\tabcolsep}{0.9mm}
  \renewcommand{\arraystretch}{0.95}
  \caption{No-reference quantitative comparisons on unpaired low-light datasets.}
  \vspace{-2mm}
  \resizebox{\linewidth}{!}{
    \begin{tabular}{l!{\vrule}cc!{\vrule}cc!{\vrule}cc!{\vrule}cc!{\vrule}cc}
    \toprule
    \multirow{2}{*}{\textbf{Methods}}
      & \multicolumn{2}{c!{\vrule}}{\textbf{DICM}}
      & \multicolumn{2}{c!{\vrule}}{\textbf{LIME}}
      & \multicolumn{2}{c!{\vrule}}{\textbf{MEF}}
      & \multicolumn{2}{c!{\vrule}}{\textbf{NPE}}
      & \multicolumn{2}{c}{\textbf{VV}} \\
      & BRISQUE $\downarrow$ & NIQE $\downarrow$
      & BRISQUE $\downarrow$ & NIQE $\downarrow$
      & BRISQUE $\downarrow$ & NIQE $\downarrow$
      & BRISQUE $\downarrow$ & NIQE $\downarrow$
      & BRISQUE $\downarrow$ & NIQE $\downarrow$ \\
    \midrule

    LIME~\cite{Dataset.LL.LIME} {\scriptsize TIP'17}
      & 23.36 & 3.812 & 18.98 & 4.085 & 16.34 & 3.558 & 17.30 & 4.196 & - & - \\

    KinD~\cite{LL.KinD} {\scriptsize ACM MM'19}
      & 29.11 & 4.139 & 25.21 & 4.762 & 27.51 & 3.875 & 18.09 & 4.165 & 23.41 & 3.027 \\

    ZeroDCE~\cite{LL.Zero-DCE} {\scriptsize CVPR'20}
      & 23.52 & 3.723 & 18.48 & 3.771 & 16.63 & \second{3.283} & 15.73 & 3.946 & - & - \\

    MIRNet~\cite{IR.MIRNet} {\scriptsize ECCV'20}
      & 24.00 & 4.738 & 21.04 & 5.005 & 28.28 & 4.549 & 24.59 & 4.556 & 26.12 & 4.084 \\

    ZeroDCE++~\cite{LL.Zero-DCE} {\scriptsize TPAMI'21}
      & 21.37 & 3.824 & 17.22 & 3.965 & 13.60 & 3.400 & 12.94 & 4.021 & - & - \\

    RUAS~\cite{LL.RUAS} {\scriptsize CVPR'21}
      & 33.44 & 5.270 & 22.78 & 4.302 & 19.27 & 3.828 & 41.21 & 5.678 & 27.68 & 4.244 \\

    LEDNet~\cite{IR.LEDNet} {\scriptsize TIP'22}
      & 19.56 & 4.077 & 21.30 & 5.059 & 20.63 & 4.194 & 14.35 & 4.875 & 19.15 & 3.109 \\

    LLFlow~\cite{LL.LLFlow} {\scriptsize AAAI'22}
      & 24.90 & 3.831 & 26.60 & 5.073 & 26.97 & 3.924 & 19.68 & 4.204 & 28.14 & 3.175 \\

    URetinexNet~\cite{LL.URetinexNet} {\scriptsize CVPR'22}
      & 24.27 & 4.152 & 23.22 & 4.240 & 21.45 & 3.791 & 24.42 & 4.693 & 23.26 & 3.003 \\

    UHDFour~\cite{LL.UHDFour} {\scriptsize ACM MM'23}
      & 33.83 & 9.941 & 37.82 & 4.506 & 30.65 & 4.270 & 19.46 & 4.501 & 138.60 & 46.117 \\

    RetinexMamba~\cite{LL.RetinexMamba} {\scriptsize ICONIP'23}
      & 16.92 & 4.189 & 19.35 & 4.240 & 19.74 & 4.217 & 18.79 & 4.232 & \first{12.72} & 3.489 \\

    URWKV~\cite{IR.URWKV} {\scriptsize CVPR'25}
      & 20.63 & 4.001 & 19.65 & 4.407 & 25.49 & 4.187 & 16.46 & 4.190 & 22.35 & 3.523 \\

    \midrule

    Retinexformer~\cite{LL.Retinexformer} {\scriptsize ICCV'23}
      & \second{15.66} & 3.973 & 14.68 & 4.167 & 14.12 & 4.363 & 15.38 & 4.691 & 16.33 & 2.989 \\

    \ours
    Retinexformer + Ours
      & 19.18 \ls{3.52} & 3.966 \gt{0.007}
      & 16.01 \ls{1.34} & 3.903 \gt{0.264}
      & 12.58 \gt{1.54} & 3.522 \gt{0.841}
      & 10.53 \gt{4.85} & 3.977 \gt{0.713}
      & 17.09 \ls{0.76} & \first{2.824} \gt{0.165} \\
    \midrule

    DarkIR~\cite{LL.DarkIR} {\scriptsize CVPR'25}
      & 16.95 & 3.791 & \second{13.32} & 3.953 & 12.13 & 3.482 & \second{10.09} & \second{3.923} & 14.88 & 2.998 \\

    \ours
    DarkIR + Ours
      & \first{15.49} \gt{1.45} & \second{3.667} \gt{0.124}
      & \first{10.00} \gt{3.32} & \first{3.652} \gt{0.301}
      & \first{11.60} \gt{0.53} & 3.463 \gt{0.020}
      & \first{9.06} \gt{1.03} & \first{3.818} \gt{0.105}
      & \second{14.14} \gt{0.75} & 2.868 \gt{0.130} \\
    \midrule

    HVI-CIDNet~\cite{LL.HVI-CIDNet} {\scriptsize CVPR'25}
      & 25.50 & \first{3.571} & 18.64 & 4.052 & 20.52 & 3.701 & 16.99 & 4.118 & 27.35 & 3.313 \\

    \ours
    HVI-CIDNet + Ours
      & 17.96 \gt{7.54} & 3.706 \ls{0.135}
      & 15.16 \gt{3.48} & \second{3.701} \gt{0.352}
      & \second{11.85} \gt{8.67} & \first{3.249} \gt{0.452}
      & 13.88 \gt{3.11} & 3.991 \gt{0.127}
      & 16.69 \gt{10.66} & \second{2.842} \gt{0.471} \\

    \bottomrule
    \end{tabular}
  }

  \vspace{1mm}
  \begin{minipage}{\linewidth}
    \footnotesize
    \raggedright
    The best and second-best results are highlighted in \first{bold} and \second{underline}, respectively.
    \gtsign{+}(\lssign{-}) denotes the improvement (drop) relative to the corresponding baseline. 
  \end{minipage}
  \label{tab:unpaired_results}
\end{table*}

%% file: tables/supp-ablation.tex
\begin{table}[t]
  \vspace{-2mm}
  \centering
  \caption{Ablation of global interaction.}
  \vspace{-3mm}
  \label{tab:abl:gamut}
  \setlength{\tabcolsep}{5pt}
  \renewcommand{\arraystretch}{1.12}
  \resizebox{\linewidth}{!}
  {
    \begin{tabular}{c!{\vrule}l!{\vrule}ccc}
    \toprule
    \rowcolor{gray}
    \# & Vertex Values Learning Strategy & PSNR \up & SSIM \up & LPIPS \down \\
    \midrule
    \circnum{1} & AdaICF~\cite{LL.HVI-CIDNet}  & 23.9560 & 0.8701 & 0.1080 \\
    \circnum{2} & 1DLUT  & 24.1666 & 0.8727 & 0.1015 \\
    \circnum{3} & 1DLUT w/ Monotonicity  & 23.7778 & 0.8734 & 0.1024 \\
    \circnum{4} & 1DLUT w/ Global Interaction & \textbf{24.2355} & \textbf{0.8746} & \textbf{0.0977} \\
    \bottomrule
    \end{tabular}
  }
  
  \vspace{1mm}
  \begin{minipage}{\linewidth}
    \footnotesize
    \raggedright
    AdaICF denotes adaptive intensity collapse function $C_k(l)=\sqrt[k]{\sin\!\left(\frac{\pi l}{2}\right)+\varepsilon}$ adopted from HVI-CIDNet~\cite{LL.HVI-CIDNet}, where $k\in\mathbb{Q}^+$ is a trainable parameter, $\epsilon=1\times10^{-8}$ is a small constant to avoid gradient explosion, and $l$ is the lightness map.
  \end{minipage}
\end{table}